\RequirePackage{fix-cm}
\documentclass[smallextended]{for_arxiv}    

\usepackage{graphicx}
\usepackage{amsmath}
\usepackage{amssymb}
\usepackage{lineno}
\usepackage{color}
\usepackage{hyperref}

\DeclareMathAlphabet{\mathitbf}{OT1}{ptm}{bx}{it} 
\newcommand{\mat}[1]{\mathitbf{#1}} 
\newcommand{\twod}{\mbox{\textsc{2-d}}} 
\newcommand{\threed}{\mbox{\textsc{3-d}}} 
\newcommand{\tp}{{\mspace{0mu}\scriptscriptstyle\top}} 
\newcommand{\inh}{{\mspace{-1mu}\vartriangle}}
\newcommand{\lf}{\ell} 
\newcommand{\rt}{r} 
\newcommand{\eqand}{\quad\text{and}\quad}
\newcommand{\eqwhere}{\quad\text{where}\quad}

\newcommand{\tof}{\mbox{\textsc{t}o\textsc{f}}}

\newcommand{\rgb}{\textsc{rgb}}
\newcommand{\tofrgb}{\mbox{\tof+{\footnotesize2}\hspace{0pt}\textsc{rgb}}}

\begin{document}

\title{\bf Cross-calibration of Time-of-flight and\\ Colour Cameras\footnote{This work has received funding from Agence Nationale de la Recherche under the MIXCAM project number ANR-13-BS02-0010-01.}}

\author{Miles Hansard \and Georgios Evangelidis \and Quentin Pelorson \and Radu Horaud}
\institute{Miles Hansard \at School of EECS, Queen Mary, University of London,\\ Mile End Road, London E1 4NS, UK.\\ \email{miles.hansard@eecs.qmul.ac.uk} \and Georgios Evangelidis \and Quentin Pelorson \and Radu Horaud \at \textsc{Inria} Grenoble Rh\^{o}ne Alpes, 655 avenue de l'Europe,\\ 38330 Montbonnot, France. \\ \email{georgios.evangelidis@inria.fr, quention.pelorson@inria.fr, radu.horaud@inria.fr}}

\maketitle

\begin{abstract}
Time-of-flight cameras provide depth information, which is complementary to the photometric appearance of the scene in ordinary images. It is desirable to merge the depth and colour information, in order to obtain a coherent scene representation. However, the individual cameras will have different viewpoints, resolutions and fields of view, which means that they must be mutually calibrated. This paper presents a geometric framework for the resulting multi-view and multi-modal calibration problem. It is shown that three-dimensional projective transformations can be used to align depth and parallax-based representations of the scene, with or without Euclidean reconstruction.
A new evaluation procedure is also developed; this allows the reprojection error to be decomposed into calibration and sensor-dependent components. 
The complete approach is demonstrated on a network of three time-of-flight and six colour cameras. The applications of such a system, to a range of automatic scene-interpretation problems, are discussed.
\end{abstract}


\section{Introduction}

The segmentation of multi-view video data, with respect to physically distinct objects of interest, is an essential task in automatic scene-interpretation. Visual segmentation can be based on colour, texture, parallax and motion information (e.g.~\cite{hoiem-2007,tron-2007}). The task remains very difficult, however, owing to the combined effects of non-rigid surfaces, variable lighting, and occlusion. It has become clear that depth cameras can make an important contribution to scene understanding, by enabling direct \emph{depth segmentation}, based on the measured scene-structure (see CVIU special issue \cite{savarese-2009}). This approach is also highly effective for dynamic tasks, such as body tracking and action recognition \cite{liu-2014}. Furthermore, if depth and colour information can be merged into a single representation, then a complete \threed\ representation is possible, in principle. This is clearly desirable, because colour and texture data are essential to many other aspects of scene-understanding, such as identification and tracking \cite{alahi-2010}.

There are two major obstacles to the construction of a complete scene representation, from a multi-modal camera network. Firstly, typical depth sensors are unable to capture \rgb\ data \cite{mesa-2010}. This means that the depth and colour cameras will have different viewpoints, and so the raw data are \emph{inconsistent}. Secondly, typical \tof\ and \rgb\ cameras have limited fields of view, and so the depth and colour data are \emph{incomplete}. This paper addresses both of these problems, by showing how to estimate the geometric relationships in a multi-view, multi-modal camera network. This task will be called \emph{cross-calibration}.

In order to constrain the problem, two practical constraints are imposed from the outset. Firstly, the system will be based on \emph{time-of-flight} (\tof) cameras, in conjunction with ordinary \rgb\ cameras. The \tof\ cameras are compact, can be properly synchronized, and are industrially specified, e.g., \cite{mesa-2010}. Secondly, a \emph{modular} network of \tof+\rgb\ units is required. This is so that individual units can be added or removed, in order to optimize the scene-coverage.

\subsection{Overview}

Time-of-flight cameras can, in principle, be geometrically calibrated by standard methods \cite{hansard-2014}. This means that each pixel records an estimate of the scene-distance (range) along the corresponding ray.
The \threed\ structure of a scene can also be reconstructed from two or more 
ordinary images, via the \emph{parallax} (e.g.\ binocular disparity) between corresponding image points.
There are many advantages to be gained by combining the range and parallax
data. Most obviously, each point in a parallax-based reconstruction can be
mapped back into the original images, from which colour and texture can 
be obtained.
Parallax-based reconstructions are, however, difficult to obtain, owing to the 
difficulty of putting the
image points into correspondence. Indeed, it may be impossible to find any 
correspondences in untextured regions.
Furthermore, if a Euclidean reconstruction is 
required, then the cameras must be calibrated. The 
accuracy of the resulting reconstruction will also tend to decrease with
the distance of the scene from the cameras \cite{verri-1986}.

The range data, on the other hand, are often corrupted by noise and surface-scattering. 
The spatial resolution of current \tof\ sensors is 
relatively low, the depth-range is limited,
and the luminance signal may be unusable for rendering and for classical image processing.
It should also be recalled that \tof\ cameras, of the type used here,
cannot be used in outdoor lighting conditions.
These considerations lead to the idea of a \emph{mixed} colour and time-of-flight 
system, as described in~\cite{lindner-2010}. Such a system could, in principle, 
be used to make high-resolution Euclidean reconstructions, including photometric
information \cite{kolb-2010,gandhi-2012}.


In order to make full use of a mixed range/parallax system, it is necessary to find 
the exact geometric relationship between the different devices. In particular,
the reprojection of the \tof\ data, into the colour images, must be obtained.
This paper is concerned with the estimation of these geometric relationships.
Specifically, the aim is to align
the range and parallax reconstructions, by a suitable \threed\ transformation.

\begin{figure}[!ht]
\centering
\includegraphics[height=.3\linewidth]{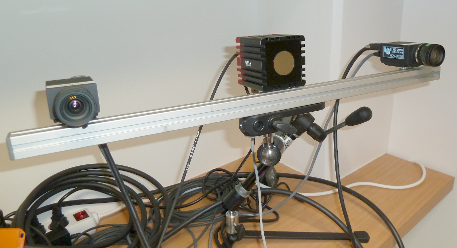}
\includegraphics[height=.3\linewidth]{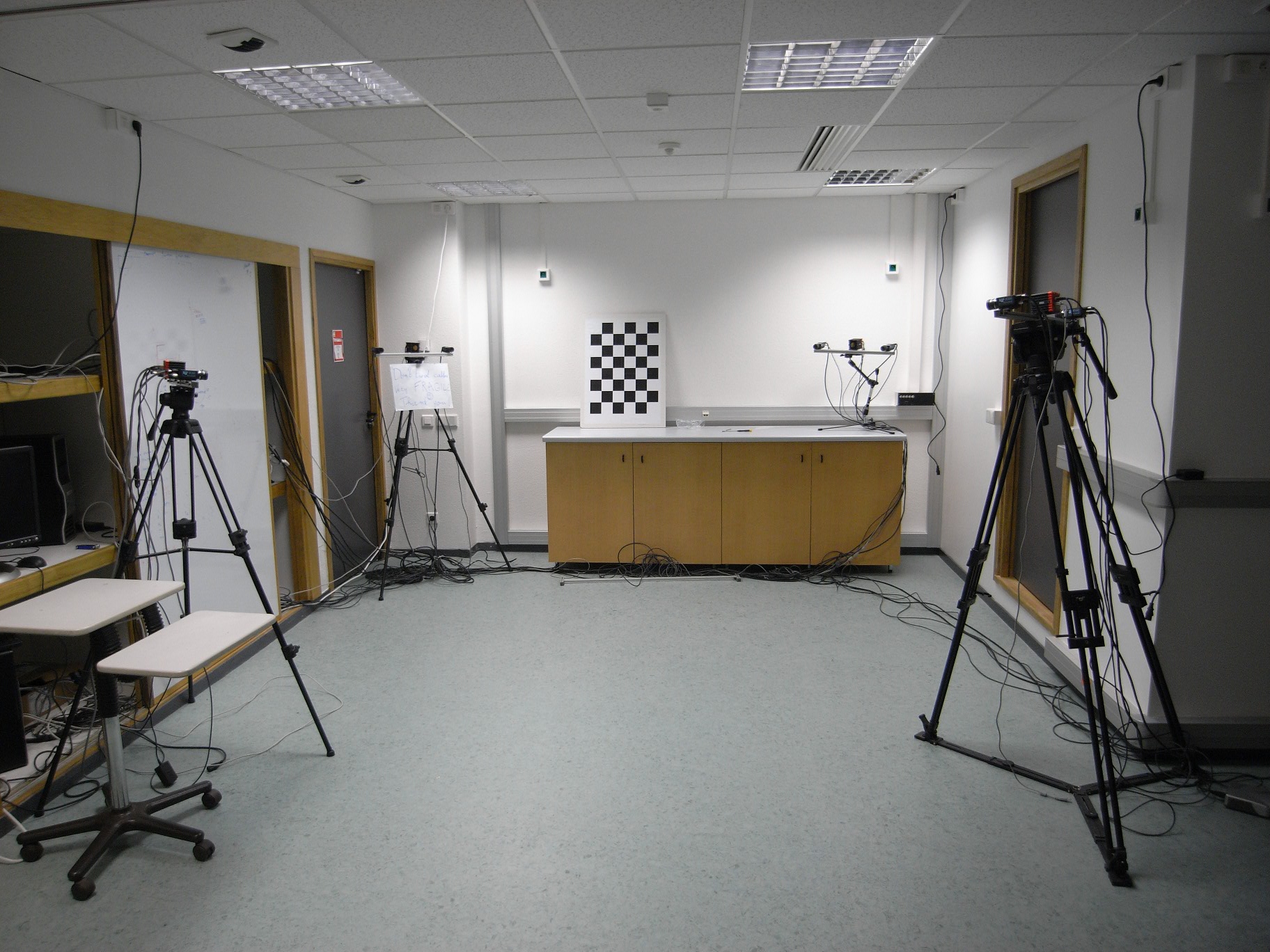}
\caption{A single \tofrgb\ system (left), comprising a time-of-flight camera in the centre, plus a pair of ordinary colour cameras. This paper addresses the problem of simultaneously cross-calibrating many such systems (right), as a foundation for scene understanding}
\label{fig:system}
\end{figure}

\subsection{Previous work}

Multi-view depth and colour camera-networks, of the kind used here, produce data-streams that are subject to a variety of geometric relationships \cite{hartley-2000}. These relationships depend on the calibration state, relative orientation, and fields of view of the different cameras. It follows that a variety of calibration strategies can be adopted. These are discussed below, with reference to the literature, and contrasted with the approach presented here.

Perhaps the simplest way to combine \rgb\ and \tof\ data is to perform an essentially \twod\ registration between the images and depth maps, as reviewed in \cite{nair-2013}; see also \cite{huhle-2007,bartczak-2009,bleiweiss-2009}. This \twod\ approach, however, can only provide an
instantaneous solution, because changes in the scene-structure produce
corresponding changes in the image-to-image mapping. Moreover, owing to the different viewpoints, a complete registration will usually be impossible. 
If the depth camera also produces a reliable intensity image, then photo-consistency can be used as a \threed\ calibration criterion. For example, Beder et al. \cite{beder-2008} (see also \cite{schiller-2008,koch-2009}) reproject the \emph{intensities} of the depth data into the colour images, and optimize the camera parameters with respect to the photo-consistency.

Zhu et al. \cite{zhu-2011} (see also \cite{zhu-2008}) present a sensor-fusion framework for the integration of \tof\ depth and binocular disparity information. This method assumes that a dense disparity map is being computed on-line, which is not required by the method presented in this paper. Furthermore, the geometric calibration method \cite{zhu-2011} requires manual identification of corresponding points, and is based on a weak perspective camera model. In contrast, our method is automatic, and is based on the more appropriate perspective camera model. However, \cite{zhu-2011} is complementary to our method, in the sense that their sensor-fusion framework (along with a dense stereo-matcher) could be combined with the projective calibration method described below. Wang and Jia \cite{wang-2012} describe a related sensor-fusion framework for Kinect (rather than \tof) depth data and colour images.

Another approach to the multi-modal calibration problem is to apply standard methods, as far as possible, to the depth cameras. Wu et al. \cite{wu-2009} describe an example of this approach. Lindner et al. \cite{lindner-2010} analyze the applicability of standard methods to \tof\ cameras, as well as characterizing the accuracy of the depth data.
Mure-Dubois and H\"{u}gli 
\cite{mure-2008} describe the Euclidean alignment of multiple \tof\ point clouds, having calibrated the cameras by standard methods.

Silva et~al.~\cite{silva-2012} describe a cross-calibration methodology that is based on the identification of \threed\ lines in the \tof\ data, which are then projected to corresponding \twod\ lines in the \rgb\ images. This method does not require a chequerboard or other calibration pattern; it does, however, require the existence and detection of straight depth-edges throughout the scene. The approach of Silva et al.~also involves a non-trivial correspondence problem, which in turn influences the calibration accuracy. Our method uses a standard chequerboard pattern, with a known number of vertices, for which the correspondence problem is relatively straightforward.

Zhang and Zhang \cite{zhang-2011} present cross-calibration methodology that is based on plane constraints, as given in \cite{zhang-1999}. This has the advantage of not requiring \twod\ features to be detected in the (low resolution) \tof\ images. However, this method cannot address the crucial issue of lens distortion, which is considerable in typical \tof\ cameras \cite{kim-2008}. 
A related Kinect-based calibration system is described by Herrera et~al.~\cite{herrera-2012}, again using the plane-based method of \cite{zhang-1999}. Herrera et al.\ give a careful analysis of the Kinect intrinsic parameters, including lens and depth distortion. 
The latter is analyzed in more detail by Teichman et~al.~\cite{teichman-2013}.
Our method does require features to be detected in the \tof\ images, but this also makes it straightforward to estimate the lens parameters, using standard techniques.

Mikhelson~et~al.~\cite{mikhelson-2013} describe an automatic method for registering a Euclidean point-cloud (obtained from a Kinect device) to its \twod\ image-projections. This method, like ours, is based on a chequerboard target. However, Mikhelson et al.~perform Euclidean \twod/\threed\ registration, in contrast to the more general projective \threed/\threed\ registration that is described below.
Finally, there are methods that perform \threed/\threed\ registration of dense data, subject to pointwise adjustments \cite{cui-2013}. This strategy can achieve very close registrations, but introduces a more complex optimization problem, which is not fully compatible with the standard calibration pipeline.

\subsection{Paper organization and contributions}

This paper is organized as follows. Section \ref{sec:parallax} briefly reviews
some standard material on projective reconstruction, while section \ref{sec:range}
describes the representation of range data in the present work. The chief
contributions of the subsequent sections are as follows:
Section~{\ref{sec:alignment}} describes a point-based
method that maps a classical multi-view reconstruction (projective or Euclidean) of the scene onto the corresponding \tof\ representation.
The data are obtained from a \tofrgb\ system, as shown in figure~\ref{fig:system}.
This does not require the colour cameras to be calibrated (although it is necessary to correct for lens distortion). It is established that this model includes a projective-linear approximation of the systematic \tof\ depth-error. 

Section~\ref{sec:multi-align} addresses the problem of multi-system alignment, which is necessary for complete scene-coverage. It is shown that this can be achieved in a way that is compatible with the individual \tofrgb\ calibrations. The complete cross-calibration pipeline, given a collection of chequerboard images, is fully automatic.

Section~\ref{sec:evaluation} contains a detailed evaluation of these methods, using several large data-sets, captured by three \tofrgb\ systems (i.e.~a nine-camera network).
In particular, section \ref{sec:cal-err} extends the usual concepts of reprojection error \cite{hartley-2000} to the multi-modal case. Section \ref{sec:tot-err} then introduces a new metric for mixed \tof/\rgb\ systems, which measures instantaneous sensor noise, as well as calibration error.
The appropriateness of the \threed\ homography transformation, as opposed to a similarity transformation, is tested in section~\ref{sec:similarity}.
Section \ref{sec:applications} discusses possible applications of these systems, including some real \threed\ reconstruction examples.
Conclusions and future directions are discussed in section~\ref{sec:conclusion}.

The system presented here is based on the approach introduced by Hansard et al.~\cite{hansard-2011,hansard-2013}. The earlier work has been improved, and extended to the case of multiple \tof\  and colour cameras. In addition, a new evaluation methodology has been developed, as described above. The automatic detection of calibration targets in the low-resolution \tof\ images, which is a pre-requisite for the methods described here, was developed in a separate paper \cite{hansard-2014}.

\section{Cross-calibration}
\label{sec:methods}

This section describes the theory of projective alignment, using the
following notation.
Bold type will be used for vectors and matrices. In particular, points 
$\mat{P}$, $\mat{Q}$ and planes $\mat{U}$, $\mat{V}$ in the \threed\ scene will 
be represented by column-vectors of homogeneous coordinates, e.g.\
\begin{equation}
\mat{P} = 
\begin{pmatrix}
\mat{P}_\inh \\
P_4
\end{pmatrix}
\eqand
\mat{U} = 
\begin{pmatrix}
\mat{U}_\inh \\
U_4
\end{pmatrix}
\end{equation}
where 
$\mat{P}_\inh = (P_1,\,P_2,\,P_3)^\tp$ and
$\mat{U}_\inh = (U_1,\,U_2,\,U_3)^\tp$. 
The homogeneous coordinates are defined up to a non-zero scaling; for example,
$\mat{P}\simeq(\mat{P}_\inh/P_4, 1)^\tp$. In particular, if $P_4=1$, then 
$\mat{P}_\inh$ contains the ordinary space coordinates of the point 
$\mat{P}$. Furthermore, if $|\mat{U}_\inh|=1$, then $U_4$ is the signed 
perpendicular distance of the plane $\mat{U}$ from the origin, and $\mat{U}_\inh$ 
is the unit normal. The point $\mat{P}$ is on the plane $\mat{U}$ if 
$\mat{U}^\tp\mspace{-2mu}\mat{P}=0$. The cross product $\mat{u}\times \mat{v}$ 
is often expressed as $(\mat{u})_\times\mat{v}$, where $(\mat{u})_\times$ is
a $3\times 3$ antisymmetric matrix. The column-vector of $N$ zeros is written
$\mat{0}_N$.
Projective cameras are represented by $3\times 4$ matrices. For example, the range
projection is
\begin{equation}
\mat{q} \simeq \mat{C}\mat{Q} \eqwhere
\mat{C} = \bigr(\mat{A}_{3\times3}\, |\, \mat{b}_{3\times 1}\bigr)
\end{equation}
is a block-decomposition of the $3\times 4$ camera matrix.
The left and right colour cameras $\mat{C}_\lf$ and $\mat{C}_\rt$ are
similarly defined, e.g.\ \mbox{$\mat{p}_\lf \simeq \mat{C}_\lf\mat{P}$}.
Table \ref{tab:notation} summarizes the geometric objects that will be aligned.

\begin{table}[!ht]
\normalsize
\begin{center}
\renewcommand\arraystretch{1}
\setlength{\tabcolsep}{2ex}
\begin{tabular}{r|ccc}
 & Observed & \multicolumn{2}{c}{Reconstructed} \\
 & Points   & Points & Planes \\ \hline \\[-2ex]
Binocular $\mat{C}_\lf$,$\mat{C}_\rt$ & $\mat{p}_\lf$, $\mat{p}_\rt$    & $\mat{P}$    & $\mat{U}$ \\[.5ex]
Range     $\mat{C}$                   & $(\mat{q},\rho)$ & $\mat{Q}$ & $\mat{V}$
\end{tabular}
\end{center}
\caption{Summary of notations in the left, right and range systems.}
\label{tab:notation}
\end{table}
Points and planes in the two systems are related by the unknown $4\times 4$ 
space-homography $\mat{H}$, so that
\begin{equation}
\mat{Q} \simeq\mat{H}\mat{P}
\eqand
\mat{V} \simeq\mat{H}^{-\tp} \mspace{-3mu}\mat{U}.
\label{eqn:homog}
\end{equation}
This model encompasses all rigid, similarity and affine transformations
in~\threed.
It preserves \emph{collinearity} and \emph{flatness}, and is linear in 
homogeneous coordinates. Note that, in the reprojection process, $\mat{H}$
can be interpreted as a modification of the camera matrices.
For example, 
\mbox{$\mat{p}_\lf\simeq \bigl(\mat{C}_\lf\mat{H}^{-1}\bigr)\mat{Q}$},
where \mbox{$\mat{H}^{-1}\mat{Q}\simeq\mat{P}$} is the point that would theoretically be reconstructed by triangulation.

The $4\times 4$ homographies $\mat{H}$ include, as special cases, the rigid transformations that would align a fully-calibrated Euclidean stereo-reconstruction to the \tof\ measurements. There are two motivations for the generalization. Firstly, it allows uncalibrated binocular reconstructions to be used, as described above. Secondly, it has been shown elsewhere \cite{kolb-2010,zhu-2011} that \tof\ data are subject to \emph{systematic} depth biases and nonlinear distortions. These are difficult to correct, owing to the lack of a complete parametric model, and to their dependence on the camera settings (e.g.\ integration time). Nonetheless, the homography model \eqref{eqn:homog} effectively includes a projective-linear approximation of the depth distortion, which is fitted along with the other transformation parameters. This model is quite powerful: it includes rational depth-distortions, with varying parameters, across each bundle of rays. 
For example, the two-parameter inverse disparity calibration, as used with Kinect devices \cite{herrera-2012}, is a special case of the homography model described here. 
Indeed, even if the \rgb\ cameras are fully calibrated, the \threed\ homographies are needed to account for depth-distortions and residual reconstruction errors, as demonstrated in~\cite{hansard-2011}. This issue will be explored in section \ref{sec:similarity}, below.

\subsection{Parallax-based reconstruction}
\label{sec:parallax}

A projective reconstruction of the scene can be obtained from matched points
$\mat{p}_{\lf k}$ and $\mat{p}_{\rt k}$, together with the fundamental 
matrix $\mat{F}$, where
\mbox{$\mat{p}_{\rt k}^\tp \mat{F}\, \mat{p}_{\lf k} = 0$}.
The fundamental matrix can be estimated automatically, using the well-established 
\textsc{ransac} method.
The camera matrices can then be determined, up to a four-parameter projective ambiguity
\cite{hartley-2000}. In particular, from 
$\mat{F}$ and the epipole $\mat{e}_{\rt}$, the cameras can be defined as
\begin{equation}
\mat{C}_\lf \simeq \bigl(\mat{I} \,|\, \mat{0}_3) 
\hspace{.6em}\text{and}\hspace{.5em}
\mat{C}_\rt \simeq 
\bigl((\mat{e}_\rt)_\times \mat{F} + 
      \mat{e}_\rt \mat{g}^\tp\, \big|\, \gamma\mat{e}_\rt\bigr).
\label{eqn:projcams}
\end{equation}
where $\gamma\ne0$ and $\mat{g}=(g_1,\,g_2,\,g_3)^\tp$ can be used to 
bring the cameras into a plausible form. This makes it easier to visualize the 
projective reconstruction and, more importantly, can improve the numerical 
conditioning of subsequent procedures. 

\subsection{Range-based reconstruction}
\label{sec:range}

The \tof\ camera $\mat{C}$ provides the range (i.e.~radial distance) $\rho$ of 
each scene-point from the camera-centre, as well as the associated 
image-coordinates $\mat{q}=(x,y,1)$.
The back-projection of this point into the scene is
\begin{equation}
\mat{Q}_\inh = 
\mat{A}^{-1}
\bigl((\rho/\alpha)\,\mat{q} - \mat{b}\bigr)
\eqwhere
\alpha = \bigl|\mat{A}^{-1} \, \mat{q}\bigr|.
\label{eqn:backproj}
\end{equation}
Hence the point $(\mat{Q}_\inh,1)^\tp$ is at distance $\rho$ from the optical centre 
$-\mat{A}^{-1}\mat{b}$, in the direction 
$\mat{A}^{-1} \mat{q}$. The scalar $\alpha$ serves to 
normalize the direction-vector. This is the standard pinhole model,
as used in \cite{beder-2007}.

The range data are noisy and incomplete, owing to illumination and scattering 
effects. This means that, given a sparse set of features in the intensity
image (of the \tof\ device), it is not advisable to use the back-projected 
point \eqref{eqn:backproj} directly. A better approach is to segment the image
of the plane in each \tof\ camera (using the the range and/or intensity data).
It is then possible to back-project \emph{all} of the enclosed points, and to 
robustly fit a plane $\mat{V}_j$ to the enclosed points $\mat{Q}_{ij}$, so that
$\mat{V}_j^\tp\! \mat{Q}_{ij} \approx 0$ if point $i$ lies on plane $j$.
Now, the back-projection
$\mat{Q}_\pi$ of each sparse feature point $\mat{q}$ can be obtained by 
intersecting the corresponding ray with the plane $\mat{V}$, so that
the new range estimate $\rho^\pi$ is
\begin{equation}
\rho^\pi = 
\frac{\mat{V}_\inh^\tp \mat{A}^{-1}\mat{b} - V_4}
{(1/\alpha)\, \mat{V}_\inh^\tp \mat{A}^{-1}\mat{q}}
\end{equation}
where $|V_4|$ is the distance of the plane to the camera centre, and
$\mat{V}_\inh$ is the unit-normal of the range plane. The new point
$\mat{Q}^\pi$ is obtained by substituting $\rho^\pi$
into \eqref{eqn:backproj}.

The choice of plane-fitting method is affected by two issues. Firstly, there
may be very severe outliers in the data, due to the photometric and geometric 
errors in the depth-estimation process. Secondly, the noise-model 
should be based on the pinhole geometry, which means that perturbations occur
radially along visual directions, which are not (in general) perpendicular
to the observed plane \cite{hebert-1992,wang-2001}. Several plane-fitting methods, 
both iterative \cite{kanazawa-1995} and non-iterative \cite{pathak-2009} have been proposed for the pinhole model. However, for \tof\ data, the chief problem is the large number of outliers. This means that a \textsc{ransac}-based method is the most effective in this context~\cite{hansard-2011}.

\subsection{Projective Alignment}
\label{sec:alignment}

It is straightforward to show that the transformation $\mat{H}$ in \eqref{eqn:homog}
could be estimated
from five binocular points $\mat{P}_k$, together with the corresponding range 
points $\mat{Q}_k$. This would provide $5\times 3$ equations, which determine 
the $4\times 4$ entries of $\mat{H}$, subject to an overall projective scaling.
It is better, however, to use the `Direct Linear Transformation' method \cite{hartley-2000}, 
which fits $\mat{H}$ to \emph{all} of the data. 
This method is based on the fact that if 
\begin{equation}
\mat{P}' = \mat{H}\mat{P}
\label{eqn:newpoint}
\end{equation}
is a perfect match for $\mat{Q}$, then $\mu\mat{Q} = \lambda\mat{P}'$, 
and the scalars $\lambda$ and $\mu$ can be eliminated between pairs of the four
implied equations \cite{csurka-1999}. This results in $\binom{4}{2}=6$ 
interdependent constraints per point. It is convenient to write these homogeneous
equations as
\begin{equation}
\begin{pmatrix}
Q_4\mat{P}_\inh' - P_4'\mspace{1mu}\mat{Q}_\inh \\[1ex]
\mat{Q}_\inh \times \mat{P}_\inh'
\end{pmatrix} =
\mat{0}_{\mspace{2mu}6}.
\label{eqn:constraints}
\end{equation}
Note that if $\mat{P}'$ and $\mat{Q}$ are normalized so 
that $P'_4=1$ and $Q_4=1$, then the magnitude of the top
half of \eqref{eqn:constraints} is simply the distance between the points.
Following F\"{o}rstner \cite{forstner-2005}, the left-hand side of 
\eqref{eqn:constraints} can be expressed as
$\bigl(\mat{Q}\bigr)_\wedge \mat{P}'$
where
\begin{equation}
\bigl(\mat{Q}\bigr)_\wedge = 
\begin{pmatrix}
Q_4 \mat{I}_3 & -\mat{Q}_\inh \\[1ex]
\bigl(\mat{Q}_\inh\bigr)_\times & \mat{0}_{\mspace{2mu}3} 
\end{pmatrix}
\label{eqn:operator}
\end{equation}
is a $6\times 4$ matrix, and 
$\bigl(\mat{Q}_\inh\bigr)_\times \mat{P}_\inh = \mat{Q}_\inh \times \mat{P}_\inh$,
as usual. The equations \eqref{eqn:constraints} can now be written in terms
of \eqref{eqn:newpoint} and \eqref{eqn:operator} as
\begin{equation}
\bigl(\mat{Q}\bigr)_\wedge \mat{H} \mat{P} = \mat{0}_{\mspace{2mu}6}.
\label{eqn:dlt-operator}
\end{equation}
This system of equations is linear in the unknown entries of $\mat{H}$, the columns
of which can be stacked into the $16\times 1$ vector $\mat{h}$.
The Kronecker product identity 
\mbox{$\mathrm{vec}(\mat{X}\,\mat{Y}\,\mat{Z}) =$} 
\mbox{$(\mat{Z}^\tp\!\otimes \mat{X})\,\mathrm{vec}(\mat{Y})$} can
now be applied, to give
\begin{equation}
\Bigl(\mat{P}^\tp \otimes \bigl(\mat{Q}\bigr)_\wedge\Bigr)
\mat{h} =  \mat{0}_{\mspace{2mu}6}
\eqwhere \mat{h} = \mathrm{vec}\bigl(\mat{H}\bigr).
\end{equation}
If $M$ points are observed on each of $N$ planes, then there are
$k=1,\ldots,MN$ observed pairs of points, $\mat{P}_k$ 
from the projective reconstruction and $\mat{Q}_k$ 
from the range back-projection.
The $MN$ corresponding $6 \times 16$ matrices 
\mbox{$\bigl(\mat{P}_k^\tp \otimes (\mat{Q}_k)_\wedge\bigr)$} are 
stacked together, to give  the complete system
\begin{equation}
\setlength\arraycolsep{0.1em} 
\begin{pmatrix}
\mat{P}_1^\tp &\otimes& \bigl(\mat{Q}_1\bigr)_\wedge \\
&\vdots& \\
\mat{P}_{MN}^\tp &\otimes& \bigl(\mat{Q}_{MN}\bigr)_\wedge
\end{pmatrix}
\mat{h} = 
\mat{0}_{\mspace{2mu}6MN}
\label{eqn:dlt-system}
\end{equation}
subject to the constraint $|\mat{h}|=1$, which excludes the trivial solution
$\mat{h} = \mat{0}_{16}$. It is straightforward to obtain an 
estimate of $\mat{h}$ from the SVD of the 
the $6MN \times 16$ matrix on the left of (\ref{eqn:dlt-system}). This solution, 
which minimizes an \emph{algebraic error} \cite{hartley-2000}, is the singular 
vector corresponding 
to the smallest singular value of the matrix. 
Note that the point coordinates should be transformed, to ensure that \eqref{eqn:dlt-system}
is numerically well-conditioned \cite{hartley-2000}. In this case the transformation
ensures that $\sum_k\mat{P}_{k\inh}=\mat{0}_3$ and 
$\frac{1}{MN}\sum_k|\mat{P}_{k\inh}|=\sqrt{3}$, where $P_{k4}=1$. The analogous 
transformation is applied to the range points $\mat{Q}_k$.

The above procedure effectively computes a projective basis in \threed, which requires five points, no four of which may be co-planar. This gives $5\times 3$ numbers, which are equivalent to the 15 parameters of the homogeneous $4\times 4$ matrix $\mat{H}$. In practice, however, we require full detection of 35 vertices per board, and around 10 boards per homography (depending on visibility constraints). Hence the procedure operates far from any degenerate cases, which would involve fewer than 15 degrees of freedom in total.

The DLT method, in practice, gives a good approximation 
$\mat{H}_{\scriptscriptstyle\mathrm{DLT}}$ of the homography \eqref{eqn:homog}. 
This can be used as 
a starting-point for the iterative minimization of a more appropriate
error measure. In particular, consider the \emph{reprojection error}
in the left image, 
\begin{equation}
E_\lf(\mat{C}_\lf) =
\sum_{k=1}^{MN}
D\bigl(\mat{C}_\lf\mat{Q}_k,\, \mat{p}_{\lf k}\bigr)^2
\label{eqn:reproj-err-points}
\end{equation}
where $D(\mat{p},\mat{q}) = |\,\mat{p}_\inh / p_3 - \mat{q}_\inh/q_3|$.
A \mbox{$12$-parameter} minimization of the reprojection error \eqref{eqn:reproj-err-points}, 
starting with the linear estimate
\mbox{$\mat{C}_\lf \leftarrow \mat{C}_\lf\mat{H}_{\scriptscriptstyle\mathrm{DLT}}^{-1}$},
is then performed by the Levenberg-Marquardt algorithm \cite{press-1992}. The 
result will be the camera matrix $\mat{C}_\lf^\star$ that best
reprojects the range data into the left image ($\mat{C}_\rt^\star$ is
similarly obtained). The solution, provided that the calibration points
adequately covered the scene volume, will remain valid for subsequent depth 
and range data.

Alternatively, it is possible to minimize the \emph{joint} reprojection error, defined as the sum
of left and right contributions,
\begin{equation}
E\bigl(\mat{H}^{-1}\bigr) = 
E_\lf\bigl(\mat{C}_\lf\mat{H}^{-1}\bigr) + 
E_\rt\bigl(\mat{C}_\rt\mat{H}^{-1}\bigr)
\label{eqn:reproj-err-points-joint}
\end{equation}
over the (inverse) homography $\mat{H}^{-1}$. The 16 parameters are again
minimized by the Levenberg-Marquardt algorithm, starting from the DLT solution
$\mat{H}_{\scriptscriptstyle\mathrm{DLT}}^{-1}$.

The difference between the separate \eqref{eqn:reproj-err-points}
and joint \eqref{eqn:reproj-err-points-joint} minimizations
is that the latter preserves the original epipolar geometry, whereas
the former does not. Recall that $\mat{C}_\lf$
$\mat{C}_\rt$, $\mat{H}$ and $\mat{F}$ are all defined up to scale, 
and that $\mat{F}$ satisfies an additional rank-two 
constraint \cite{hartley-2000}. Hence the underlying parameters can be counted
as \mbox{$(12-1)+(12-1)=22$} in the separate minimizations, and as
\mbox{$(16-1)=15$} in the joint minimization. The fixed epipolar geometry 
accounts for the $(9-2)$ missing parameters in the joint minimization.
If $\mat{F}$ is known to be very accurate (or must be preserved) then 
the joint minimization \eqref{eqn:reproj-err-points-joint} should be performed.
This will also preserve the original binocular triangulation, provided
that a projective-invariant method was used \cite{hartley-1997}.
However, if minimal reprojection error is the objective, then the cameras
should be treated separately. This will lead to a new fundamental matrix
\mbox{$\mat{F}^\star = (\mat{e}_\rt^\star)_\times\mat{C}_\rt^\star(\mat{C}_\lf^\star)^+$},
where $(\cdot)^+$ is the generalized inverse, and
$\mat{C}_\lf^\star$,
$\mat{C}_\rt^\star$ are the optimized camera-matrices. The epipole in the right-hand image is obtained from \mbox{$\mat{e}_\rt^\star = \mat{C}_\rt^\star\mat{d}_\lf^\star$},
where the vector $\mat{d}_\lf^\star$ represents the nullspace 
\mbox{$\mat{C}_\lf^\star\mat{d}_\lf^\star=\mat{0}_3$}.

\section{Multi-System Alignment}
\label{sec:multi-align}

The methods described in section \ref{sec:alignment} can be used to calibrate a single \tofrgb\ system; the joint calibration of \emph{several} such systems
will now be explained. 
In this section the notation $\mat{P}_i$ will be used for the binocular coordinates (with 
respect to the left camera) of a point in the \mbox{$i$-th} system, and likewise 
$\mat{Q}_i$ for the \tof\ coordinates of a point in the same system.
Hence the \mbox{$i$-th} \tof, left and right \rgb\ cameras (sharing the same physical mounting) have the form
\begin{equation}
\begin{gathered}
\mat{C}_i \simeq \bigl(\mat{A}_i \,|\, \mat{0}_3)\\
\mat{C}_{\lf i} \simeq \bigl(\mat{A}_{\lf i} \,|\, \mat{0}_3)
\quad\text{and}\quad 
\mat{C}_{\rt i} \simeq \bigl(\mat{A}_{\rt i} \,|\, \mat{b}_{\rt i})
\end{gathered}
\label{eqn:stereo-cams}
\end{equation}
where $\mat{A}_i$ and $\mat{A}_{\lf i}$ contain only \emph{intrinsic} 
parameters, whereas $\mat{A}_{\rt i}$ also encodes the relative orientation 
of $\mat{C}_{\rt i}$ with respect to $\mat{C}_{\lf i}$.
Each system has a transformation $\mat{H}_i^{-1}$ that maps \tof\ points
$\mat{Q}_i$ into the corresponding \rgb\ coordinate system of $\mat{C}_{\lf i}$.
Furthermore, let the $4\times 4$
matrix $\mat{G}_{ij}$ be the transformation from system $j$, mapping \emph{back} to system $i$ (between different physical mountings).
This matrix, in the binocularly-calibrated case, is a rigid \threed\ transformation.
However, by analogy with the \tof-to-\rgb\ matrices $\mat{H}$, each $\mat{G}_{ij}$ could be a \emph{projective} transformation in the uncalibrated case; this would allow Euclidean structure to propagate from the \tof\ measurements, across the entire camera network.

\begin{figure}[!ht]
\centering
\includegraphics[width=0.75\linewidth]{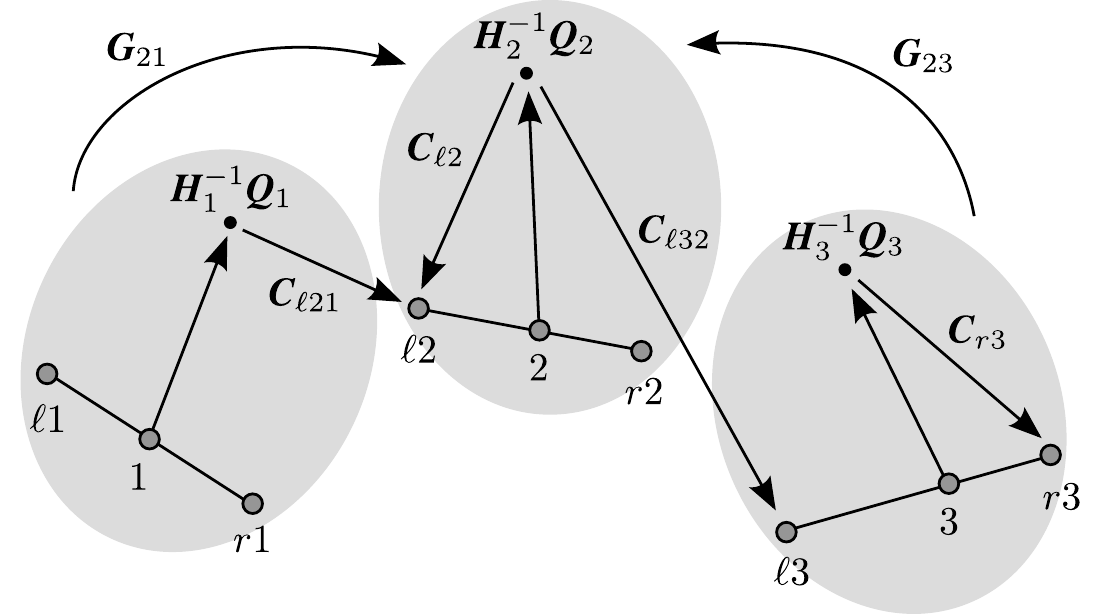}
\caption{Example of a three \tofrgb\ setup, with \tof\ cameras labelled 1,2,3; cf.~fig.~\ref{fig:system}.
Each ellipse represents a separate system, with system 2 chosen as the reference.
The arrows (with camera-labels) show some possible \tof-to-\rgb\ projections.
For example, a point $\mat{P}_2\simeq\mat{H}_2^{-1}\mat{Q}_2$ in the centre
projects directly to \rgb\ view $\ell2$ via $\mat{C}_{\lf 2}$, whereas the same point 
projects to $\lf3$ via $\mat{C}_{\lf 32} = \mat{C}_{\lf 3} \mat{G}_{32}$.}
\label{fig:cams}
\end{figure}

The left and right cameras, in all cases, that project a scene-point $\mat{P}_{\!j}$
in coordinate \mbox{system $j$} to image-points $\mat{p}_{\lf i}$ and $\mat{p}_{\rt i}$ 
in system~$i$ are 
\begin{equation}
\mat{C}_{\lf ij} = 
\mat{C}_{\lf i} \,\, \mat{G}_{ij} \quad\text{and}\quad 
\mat{C}_{\rt ij} = 
\mat{C}_{\rt i} \,\, \mat{G}_{ij}.
\label{eqn:cams-ij}
\end{equation}
Note that if a single global coordinate system is chosen to coincide with the \mbox{$k$-th} \rgb\ system,
then a point $\mat{P}_k$ projects via
$\mat{C}_{\lf ik}$
and
$\mat{C}_{\rt ik}$.
These two cameras are respectively equal to $\mat{C}_{\lf i}$ and $\mat{C}_{\rt i}$ 
in (\ref{eqn:stereo-cams}) only when $i=k$, such that $\mat{G}_{ij}=\mat{I}$ in~(\ref{eqn:cams-ij}).
A typical three-system configuration is shown in fig.~\ref{fig:cams}. 

The transformation $\mat{G}_{ij}$ can only be estimated directly if there is a region
of common visibility between systems $i$ and $j$. If this is not the case (as
when the systems face each other, such that the front of the calibration
board is not simultaneously visible), then $\mat{G}_{ij}$ can be computed 
indirectly. For example, 
$\mat{G}_{02} = \mat{G}_{01} \, \mat{G}_{12}$
where
$\mat{P}_2 = \mat{G}_{12}^{-1} \mat{G}_{01}^{-1} \mat{P}_0$. Note that, in all cases, the stereo-reconstructed
points $\mat{P}$ are used to estimate these transformations This is because they are always more reliable than the \tof\ points $\mat{Q}$, as demonstrated below.

\section{Evaluation}
\label{sec:evaluation}

The following sections will describe the accuracy of a nine-camera setup, 
calibrated by the methods described above. Section \ref{sec:cal-err} will evaluate
\emph{calibration} error, whereas section \ref{sec:tot-err} will evaluate 
\emph{total} error. The former is essentially a fixed function of the estimated 
camera matrices, for a given scene. The latter also includes the range-noise 
from the \tof\ cameras, which varies from moment to moment (due to intrinsic noise, changing illumination, and object motion). The importance of
this distinction will be discussed.
In section \ref{sec:similarity} we analyze the avantages of using homographies to align the data, as opposed to similarity transformations. Finally, in section \ref{sec:applications} we demonstrate some applications of the complete system.

The setup consists of three rail-mounted \tofrgb\ systems, $i=1\ldots 3$, as in 
fig.~\ref{fig:cams}. 
The stereo baselines are 17cm on average, and the \tof\ cameras are separated 
by 107cm on average. The \rgb\ images are $1624\times 1224$, whereas the Mesa
Imaging \mbox{\textsc{sr}{\footnotesize4000}}~\tof\ images are $176\times 144$, with a depth range of 500cm \cite{mesa-2010}. 
The three stereo systems are first
calibrated by standard methods~\cite{hartley-2000}, returning a full Euclidean decomposition of
$\mat{C}_{\lf i}$ and $\mat{C}_{\rt i}$, as well as the associated
lens parameters.
The lenses of the \tof\ cameras are also calibrated by standard methods \cite{hartley-2000}.
This removes some of the radial depth deformation that has been observed in the \tof\ data \cite{cui-2013}.  
The matrices $\mat{G}_{ij}$ are rigid-body transformations, which are estimated by the SVD method of Arun et al.~\cite{arun-1987}. 
Projective alignment is preferred for the \tof/\rgb\ alignment, for reasons discussed in sections~\ref{sec:methods} and \ref{fig:hom-sim}. Hence the transformations $\mat{H}_j^{-1}$ are $4\times 4$ homographies, estimated by the method of section \ref{sec:alignment}. Specifically, the DLT solutions were refined by Levenberg-Marquardt minimization of the joint geometric error, as in~(\ref{eqn:reproj-err-points-joint}).

\subsection{Calibration Error}
\label{sec:cal-err}

The calibration error is measured by first taking \tof\ points $\mat{Q}^\pi_j$ corresponding
to chequerboard \emph{vertices} on the reconstructed calibration plane $\pi_j$ in system $j$, as described
in section~\ref{sec:range}. These can then be
projected into a pair of \rgb\ images in system $i$, so that the geometric image-error
$E^{\mathrm{cal}}_{ij} = \frac{1}{2}\bigl(E^{\mathrm{cal}}_{\lf ij} + E^{\mathrm{cal}}_{\rt ij}\bigr)$ can be 
computed, where
\begin{equation}
E^{\mathrm{cal}}_{\lf ij} = 
\frac{1}{|\pi|}
\sum_{\mat{Q}_j^\pi}
D\Bigl( \mat{C}_{\lf ij}\, \mat{H}_j^{-1}\mat{Q}_j^\pi, \,\,\, \mat{p}_{\lf i} \Bigr)
\label{eqn:cal-err}
\end{equation}
and $E^{\mathrm{cal}}_{\rt ij}$ is similarly defined. 
The function $D(\cdot,\cdot)$ computes the image-distance between two inhomogenized points,
as in (\ref{eqn:reproj-err-points}), and the denominator corresponds to the number of vertices
on the board, with $|\pi|=35$ in the present experiments. 
The measure~(\ref{eqn:cal-err}) can of course be averaged over all images in which the board is visible.
The \rgb\ cameras were calibrated, by standard methods, as described above. Subpixel accuracy was obtained, which confirms the accuracy of the camera matrices $\mat{C}_{\lf i}$ and $\mat{C}_{\rt i}$, as well as the inter-system matrices $\mat{G}_{ij}$, where $i,j=1,2,3$. 

\begin{figure*}[!t]
\centering
\includegraphics[width=1\linewidth]{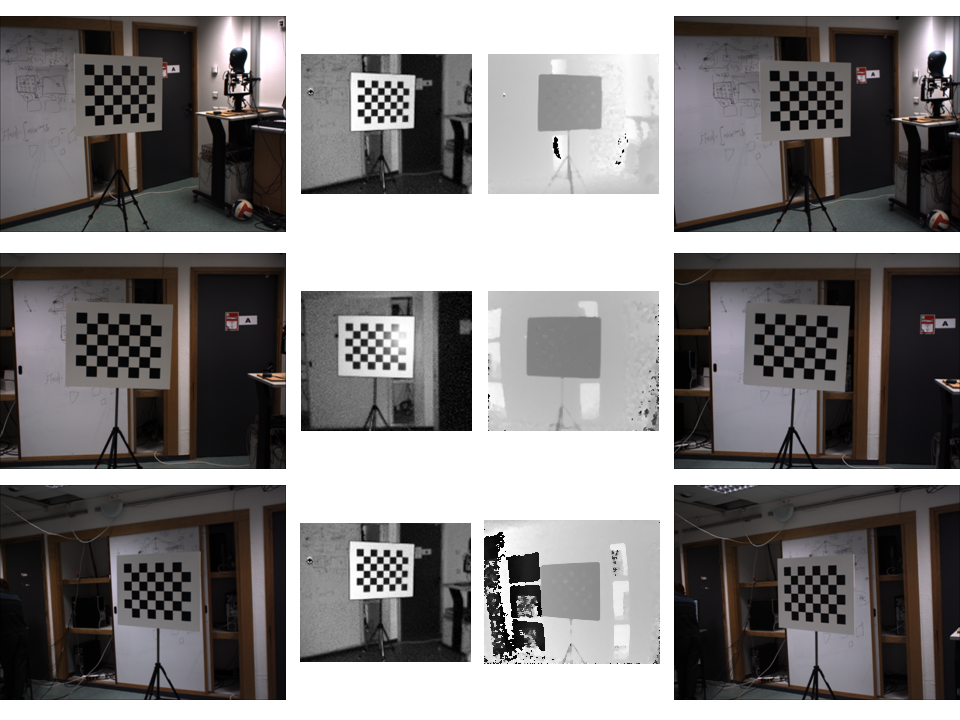}
\caption{Calibration images from synchronized captures. Each row corresponds to the image triplet of a \tofrgb\ system, while the range images are shown as well. Grayscale images of \tof\ sensors have undergone lens undistortion and contrast enhancement.}
\label{fig:camera_configuration}
\end{figure*}

\begin{figure*}[!ht]
\centering
\includegraphics[width=1\linewidth,trim=0mm 5mm 0mm 0mm,clip]{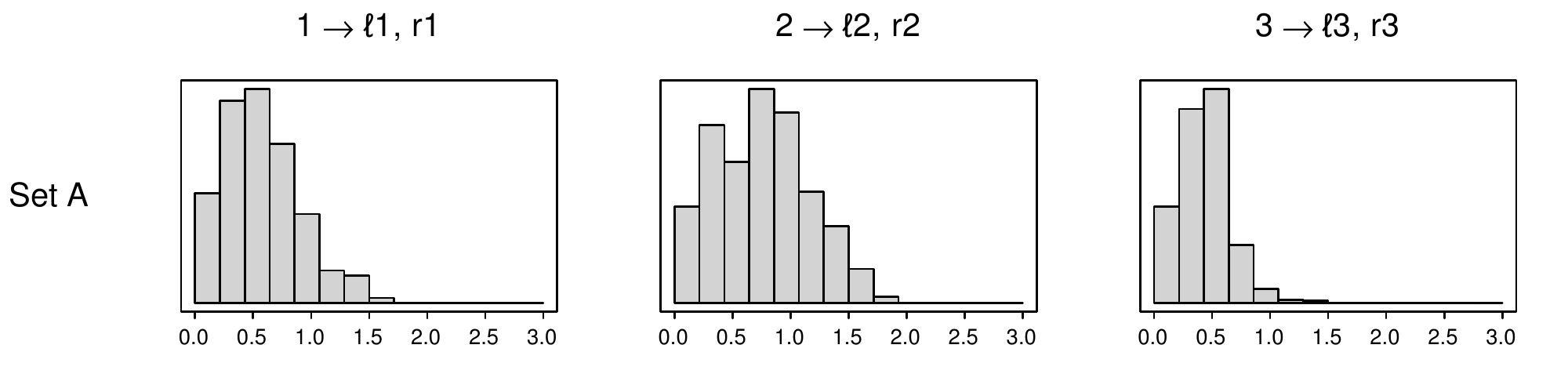}\\
\includegraphics[width=1\linewidth,trim=0mm 5mm 0mm 5mm,clip]{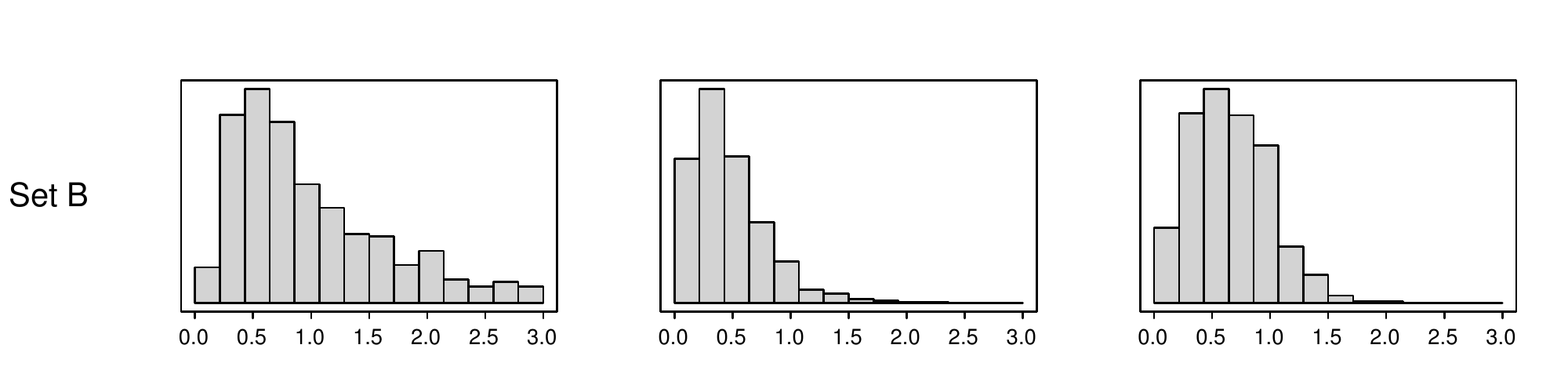}\\
\includegraphics[width=1\linewidth,trim=0mm 0mm 0mm 5mm,clip]{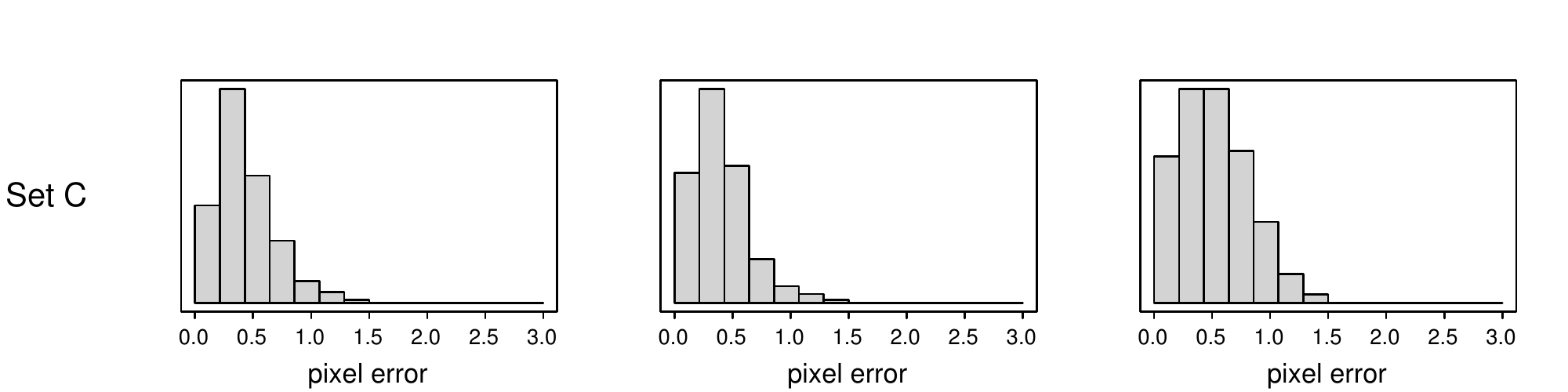}
\caption{Calibration error (\ref{eqn:cal-err}), measured by projecting the fitted \tof\ points $\mat{Q}^\pi_i$ to the left and right \rgb\ images ($1624\times 1224$) of the \emph{respective} systems $i=1,2,3$. The experiment was repeated three times (A--C), with different camera configurations. Each histogram contains 1470 points, as in table \ref{tab:vts-intra}.}
\label{fig:vts-intra}
\end{figure*}

For the purpose of evaluation, a \emph{new} set of \tof-vertices $\mat{Q}^\pi_i$ were reconstructed, fitted and reprojected within each system $i$. 
This evaluation effectively tests the quality of the $\mat{H}_i^{-1}$ matrices,
by comparing $\mat{C}_{\lf i}\mat{H}_i^{-1}\mat{Q}^\pi_i$ to $\mat{p}_{\lf i}$, and analogously to points $\mat{p}_{\rt i}$ in the other image.
Note that all camera and transformations parameters are now fixed; no optimization was performed with respect to the evaluation data.
The whole experiment was performed on three large data-sets, from different capture-sessions, and with different camera configurations, labelled {A}, {B} and {C}. 
Such a configuration leads to one triplet per system as shown in fig.~\ref{fig:camera_configuration}. While an image can be fronto-parallel for one frame, it may appear very slanted to the other frames. The example of fig.~\ref{fig:camera_configuration} shows a calibration image that is almost fronto-parallel to the \tof\ frame of the central \tofrgb\ unit. 
Table \ref{tab:vts-intra} shows that the average reprojection error remains subpixel in all three data-sets. The corresponding error-distributions are shown as histograms in fig.~\ref{fig:vts-intra}.

\begin{table}[!ht]
\centering
\renewcommand{\arraystretch}{1.25}
\begin{tabular}{|c|c|c|c|c|}
\hline
Set & Mean & Median & Max & Count\\
\hline
A & 0.59 & 0.52 & 1.82 & 1470 \\
B & 0.72 & 0.58 & 4.86 & 1470 \\
C & 0.45 & 0.40 & 1.48 & 1470 \\
\hline
\end{tabular}
\vspace{1ex}
\caption{Calibration error (\ref{eqn:cal-err}), measured by projecting the fitted \tof\ vertices $\mat{Q}^\pi_i$ to the left and right \rgb\ images ($1624\times 1224$) of the \emph{respective} systems $i=1,2,3$. The experiment was repeated three times (A--C), with different camera configurations. Each statistic was computed from the left and right \rgb-reprojections of 35 vertices in 7 views of the board (total number of \twod\ points, per data-set: $3\times 2 \times 35\times 7 = 1470$).}
\label{tab:vts-intra}
\end{table}

\begin{figure}[!ht]
\centering
\includegraphics[width=.6\linewidth,trim=0mm 5mm 0mm 0mm,clip]{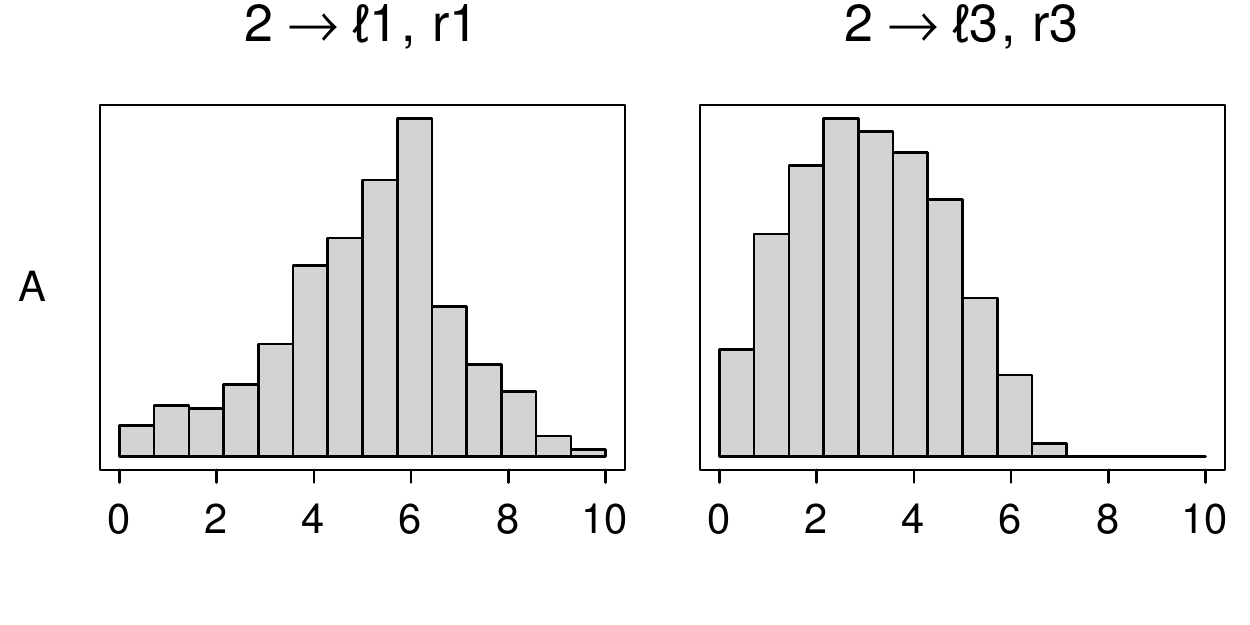}\\
\includegraphics[width=.6\linewidth,trim=0mm 5mm 0mm 5mm,clip]{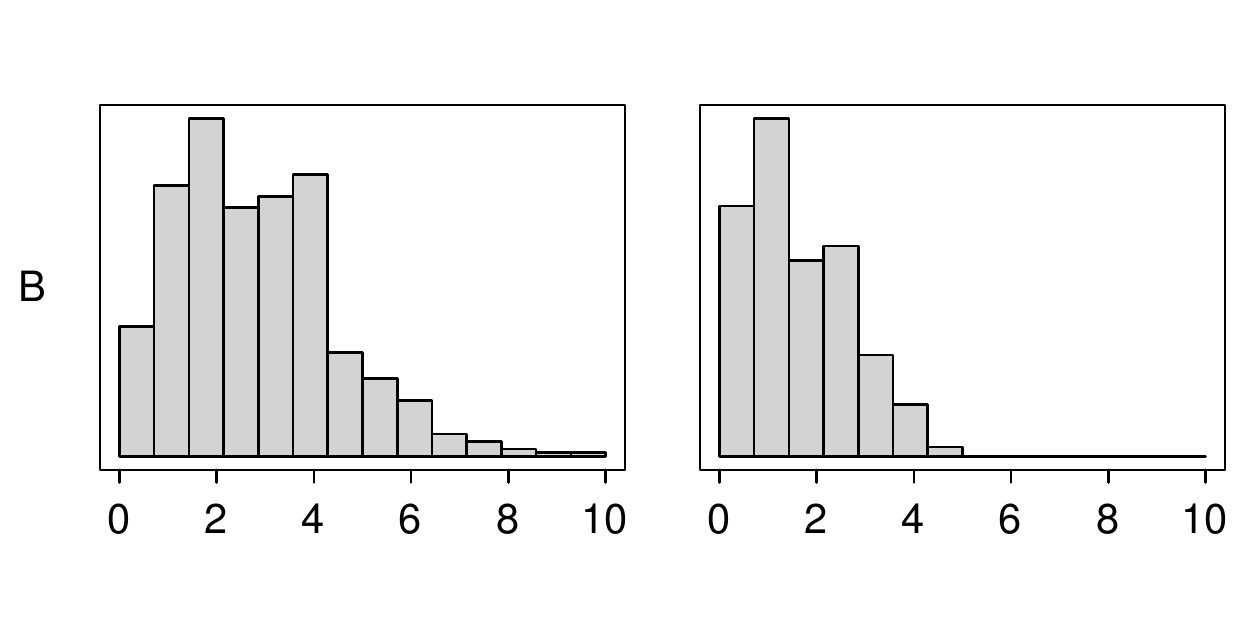}\\
\includegraphics[width=.6\linewidth,trim=0mm 0mm 0mm 5mm,clip]{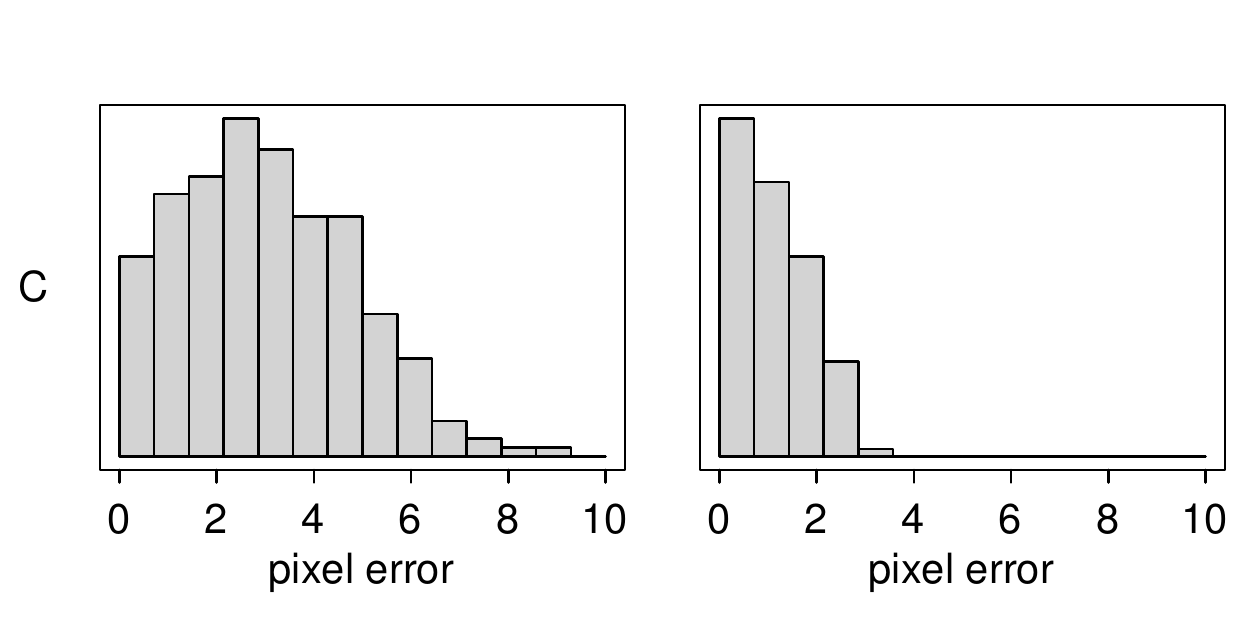}\\
\caption{Calibration error (\ref{eqn:cal-err}), measured by projecting the fitted \tof\ points $\mat{Q}^\pi_2$ to the left and right \rgb\ images ($1624\times 1224$) of two \emph{different} systems, $i=1,3$. 
Each histogram combines left-camera and right-camera measurements from 7 views of the calibration board.}
\label{fig:vts-inter}
\end{figure}

It is also interesting to consider how the above \tof-vertices reproject into \emph{different} systems. 
The error \threed\ error distribution, for a given pixel (either \tof\ or stereo) is highly anisotropic; the \emph{direction} of the corresponding ray is much more reliable than the \emph{distance} along it \cite{cui-2013}.
In practice, the different \tofrgb\ systems are distributed around the edge of the room, looking inwards. Hence a given system is likely to be seen `from the side' by at least one other system. This means that any large depth errors in the first system will not cancel-out in the re-projection to the other systems. This effect is seen clearly in figure~\ref{fig:vts-inter}, which shows calibration errors of up to several pixels, from one system to the others.

\subsection{Total Error}
\label{sec:tot-err}

The calibration error, as reported in the preceding section, is the natural 
way to evaluate the estimated cameras and homographies. It is not, however,
truly representative of the `live' performance of the complete setup. This 
is because the calibration error uses each estimated plane $\pi_j$ to replace
all vertices $\mat{Q}_j$ with the \emph{fitted} versions $\mat{Q}_j^\pi$. 
In general, however, no surface model is available, and so the raw points 
$\mat{Q}_j$ must be used as input to segmentation, meshing and rendering processes.

The total error, which combines the calibration and range errors, 
can be measured as follows. 
The $i$-th \rgb\ views of plane $\pi_j$ are related to the \tof\ image-points
$\mat{q}_j$ by the \twod\ \emph{transfer} homographies
$\mat{T}_{\lf ij}$ and $\mat{T}_{\rt ij}$, where
\begin{equation}
\mat{p}_{\lf i} \simeq \mat{T}_{\lf ij}\, \mat{q}_j \quad\text{and}\quad
\mat{p}_{\rt i} \simeq \mat{T}_{\rt ij}\, \mat{q}_j.
\label{eqn:2d-transfer}
\end{equation}
These $3\times 3$ matrices can be estimated to subpixel accuracy, by using the DLT algorithm~\cite{hartley-2000} to obtain initial homographies, and then applying area-based alignment~\cite{Evangelidis2008} to produce the final estimates
$\mat{T}_{\lf ij}$ and $\mat{T}_{\rt ij}$.

\begin{figure}[!ht]
\centering
\includegraphics[width=.8\linewidth]{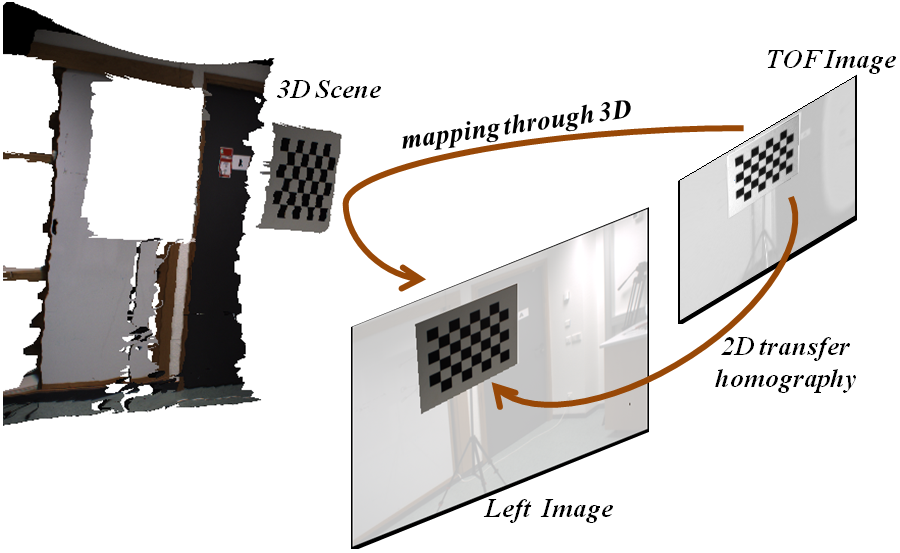}
\caption{
The \twod\ transfer homography $\mat{T}_\lf$ describes the chessboard mapping from the \tof\ frame to the left frame. In principle, this mapping should be equivalent to a mapping via the range data, i.e.\ from the \tof\ frame to the measured \threed\ plane, and back to the left frame.}
\label{fig:3d_vs_2d_mapping}
\end{figure}

Let $\Pi_j$ be the \twod\ hull (i.e.\ bounding-polygon) of
plane $\pi_j$ as it appears in the \tof\ image. Any pixel $\mat{q}_j$ in the hull (including the
original calibration vertices) can now be \emph{re-projected}
to the $i$-th \rgb\ views via the \threed\ point $\mat{Q}_j$, or 
\emph{transferred} directly by 
$\mat{T}_{\lf ij}$ and $\mat{T}_{\rt ij}$ in (\ref{eqn:2d-transfer}), 
as shown in figure~\ref{fig:3d_vs_2d_mapping}. In other words, we are able to isolate the image-to-image error that is incurred by mapping \emph{via} the \threed\ reconstruction, in relation to the direct \twod\ to \twod\ mapping defined by $\mat{T}_{\lf ij}$ and $\mat{T}_{\rt ij}$.

The total error is the average difference between the reprojections and the transfers,
$E^{\mathrm{tot}}_{ij} = \frac{1}{2}\bigl(E^{\mathrm{tot}}_{\lf ij} + E^{\mathrm{tot}}_{\rt ij}\bigr)$,
where
\begin{equation}
E^{\mathrm{tot}}_{\lf ij} = 
\frac{1}{|\Pi_j|}
\sum_{\mat{q}_j \in \Pi_j}
D\Bigl( \mat{C}_{\lf ij}\, \mat{H}_j^{-1}\mat{Q}_j, \,\,\, \mat{T}_{\lf ij}\, \mat{q}_j \Bigr)
\label{eqn:tot-err}
\end{equation}
and $E^{\mathrm{tot}}_{\rt ij}$ is similarly defined.
The view-dependent denominator $|\Pi_j|\gg|\pi|$ is the number of \emph{pixels} in the hull $\Pi_j$. Hence $E^{\mathrm{tot}}_{ij}$ is the total error, including both calibration errors and range-noise,
of \tof\ plane $\pi_j$ as it appears in the \mbox{$i$-th} \rgb\ cameras.

\begin{figure*}[!ht]
\centering
\includegraphics[width=1\linewidth,trim=0mm 5mm 0mm 0mm,clip]{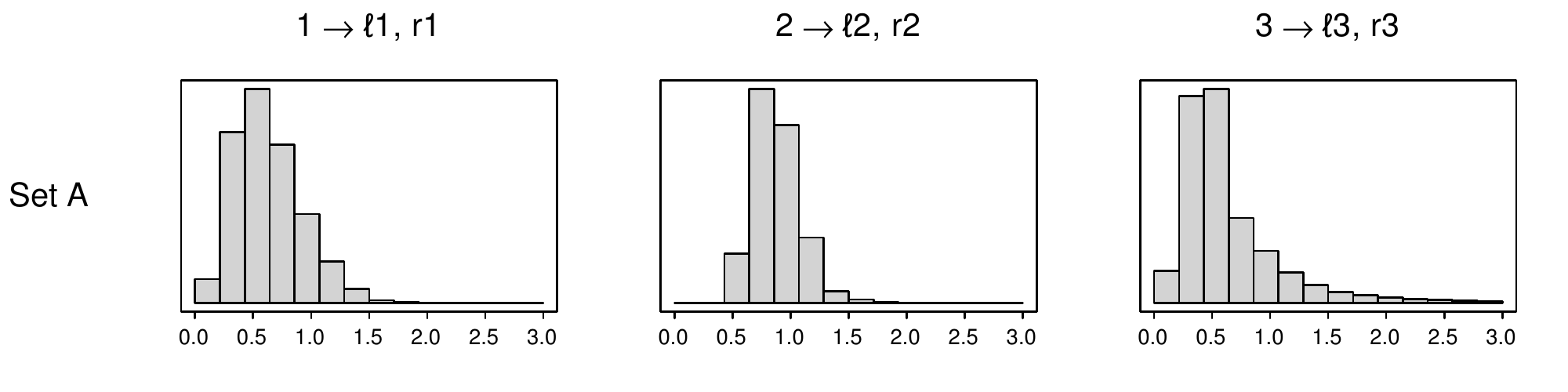}\\
\includegraphics[width=1\linewidth,trim=0mm 5mm 0mm 5mm,clip]{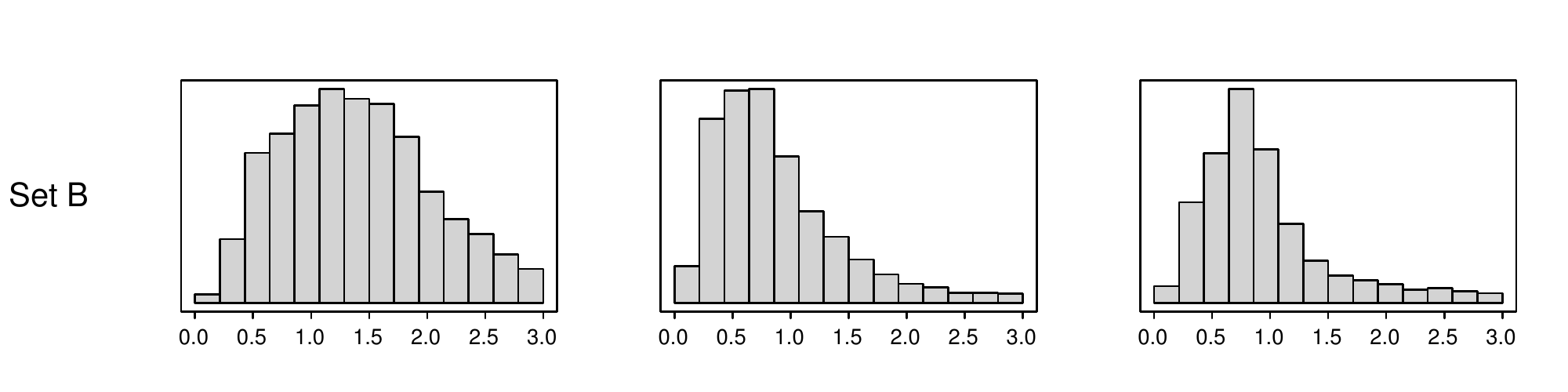}\\
\includegraphics[width=1\linewidth,trim=0mm 0mm 0mm 5mm,clip]{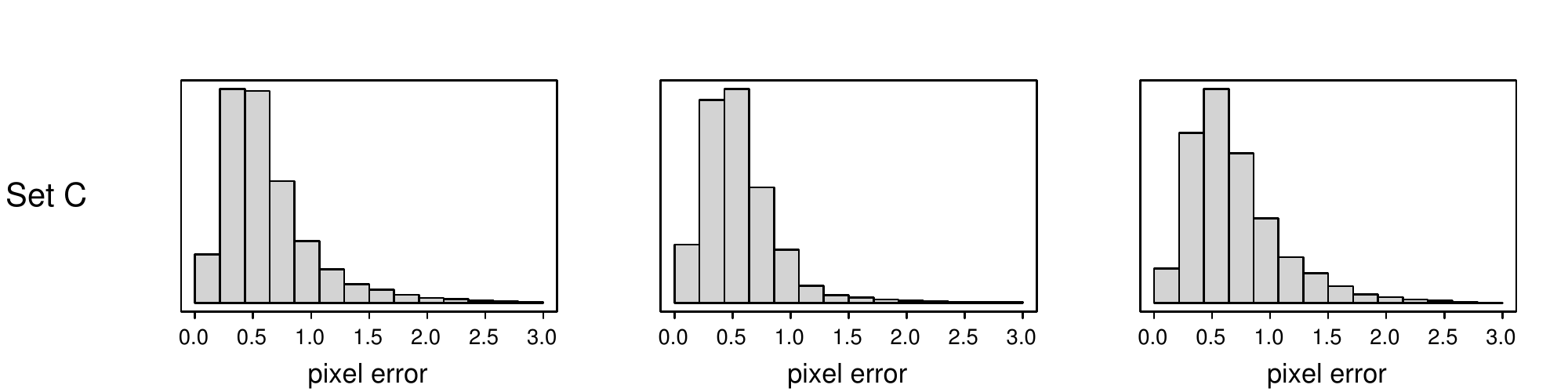}
\caption{Total error (\ref{eqn:tot-err}), measured by projecting the raw \tof\ points $\mat{Q}_i$
to the left and right \rgb\ images ($1624\times 1224$) of the \emph{respective} systems $i=1,2,3$.  
These distributions have longer and heavier tails than those of the corresponding calibration 
errors, shown in fig.~\ref{fig:vts-intra}.}
\label{fig:pix-intra}
\end{figure*}

If the \rgb\ cameras are not too far from the \tof\ camera, then the range errors tend
to be cancelled in the reprojection. This is evident from table \ref{tab:pix-intra}, in
which the total errors, on average, are not much greater than the calibration errors in 
table \ref{tab:vts-intra}. It is, however, clear from fig.~\ref{fig:pix-intra}, 
that the tails of the total error distributions are greatly increased by outliers in
the \tof\ data-stream. Although errors up to $3$ pixels are shown in distributions for the sake of clarity, maximum errors of more than five pixels are common (see table~\ref{tab:pix-intra}).

\begin{table}[!ht]
\centering
\renewcommand{\arraystretch}{1.5}
\begin{tabular}{|c|c|c|c|c|}
\hline
Set & Mean & Median & Max & Count\\
\hline
A & 0.74 & 0.71 & 6.07  & 39859 \\
B & 1.48 & 0.98 & 20.69 & 32159 \\
C & 0.63 & 0.54 & 6.51  & 29442 \\
\hline
\end{tabular}
\vspace{1ex}
\caption{Total error (\ref{eqn:tot-err}), measured by projecting the raw \tof\ points $\mat{Q}_i$
to the left and right \rgb\ images ($1624\times 1224$) of the \emph{respective} systems $i=1,2,3$. These figures characterize the raw data that is produced by the live system. This data includes all outliers in the \tof\ data-stream, which gives rise to at least one error of 20 pixels here.}
\label{tab:pix-intra}
\end{table}

\begin{figure}[!ht]
\centering
\includegraphics[width=.6\linewidth,trim=0mm 5mm 0mm 0mm,clip]{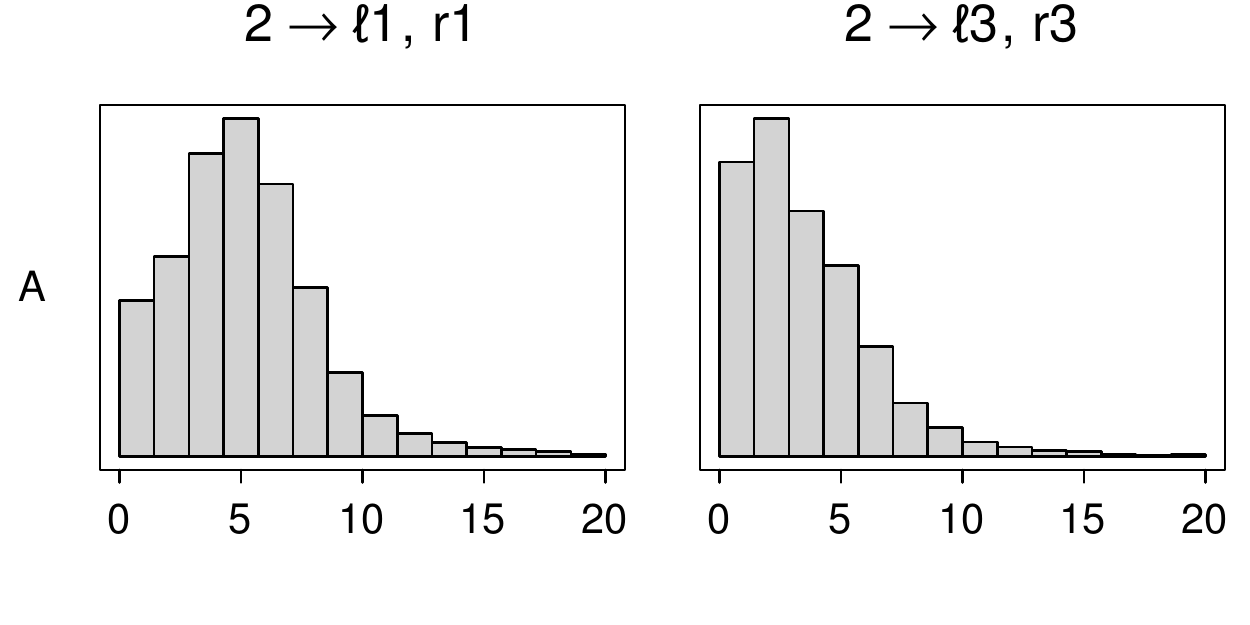}\\
\includegraphics[width=.6\linewidth,trim=0mm 5mm 0mm 5mm,clip]{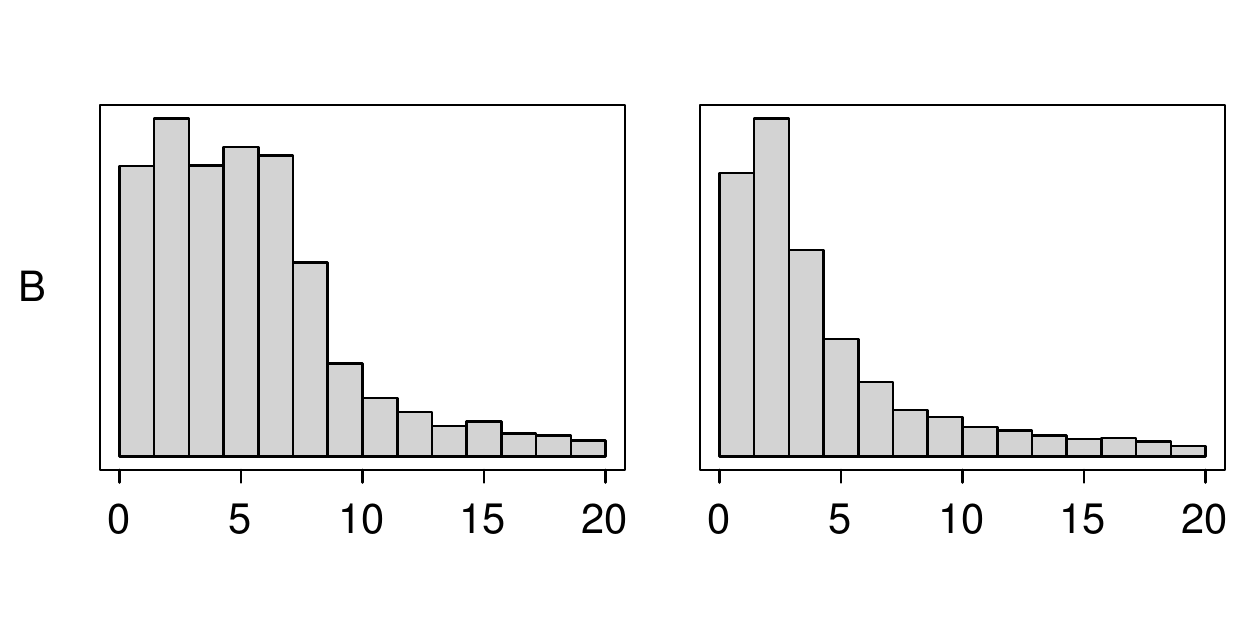}\\
\includegraphics[width=.6\linewidth,trim=0mm 0mm 0mm 5mm,clip]{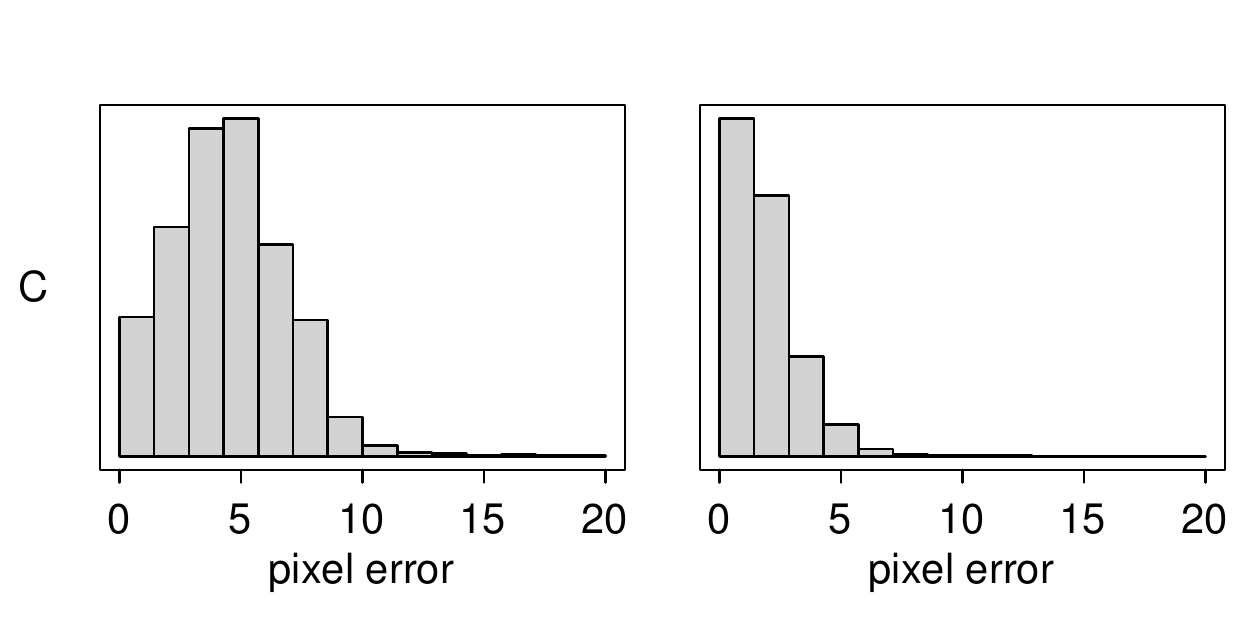}\\
\caption{Total error (\ref{eqn:tot-err}), measured by projecting the fitted \tof\ points $\mat{Q}^\pi_2$ to the left and right \rgb\ images ($1624\times 1224$) of two \emph{different} systems, $i=1,3$.
The range errors are emphasized by
the difference in viewpoints between the 
two systems. Average error is now around five pixels, 
and the noisiest \tof\ points reproject with tens of pixels of error.}
\label{fig:pix-inter}
\end{figure}

When the raw \tof\ points are reprojected to a \emph{different} system, a high total error is expected, because the sensor's range-noise may be viewed `from the side'. Indeed, fig.~\ref{fig:pix-inter} shows that a substantial 
proportion of the \tof\ points reproject with total errors in 
excess of ten pixels. These kind of gross errors are characteristic of the \tof\ data, owing to the inevitable presence of absorbing and scattering surfaces in a typical scene.

In fact, calibration error and total error are both influenced by surface orientation.
For example, the higher errors in Set~B of the data (see tables \ref{tab:vts-intra} and \ref{tab:pix-intra}) are due to the presence of some very slanted boards, in both the fitting and evaluation data-sets; this causes two problems, as follows. 
Firstly, the images of these boards are very foreshortened in some systems. This leads to increased \emph{calibration} error, because the vertices are hard to detect in the corresponding \tof\ images. 
Secondly, the strength of the reflected IR signal is reduced whenever the surface is oblique to a given \tof\ camera. If the surface is also absorbent (like the black squares of the board), then the \emph{total} error is greatly increased, due to the combined effects of scattering and absorption.

\begin{figure}[!ht]
\centering
\includegraphics[width=.7\linewidth]{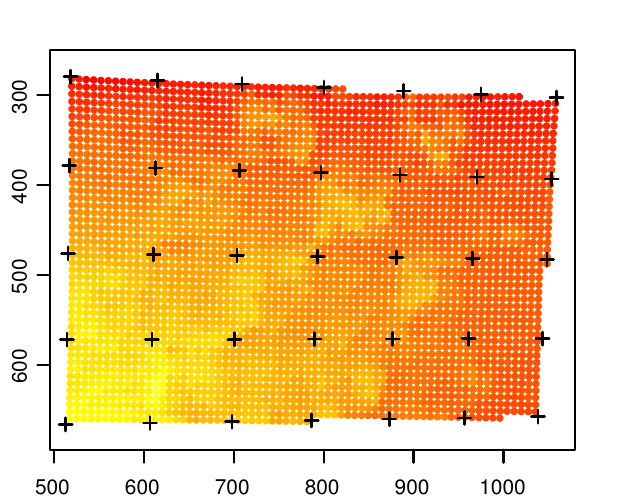}
\includegraphics[width=.7\linewidth]{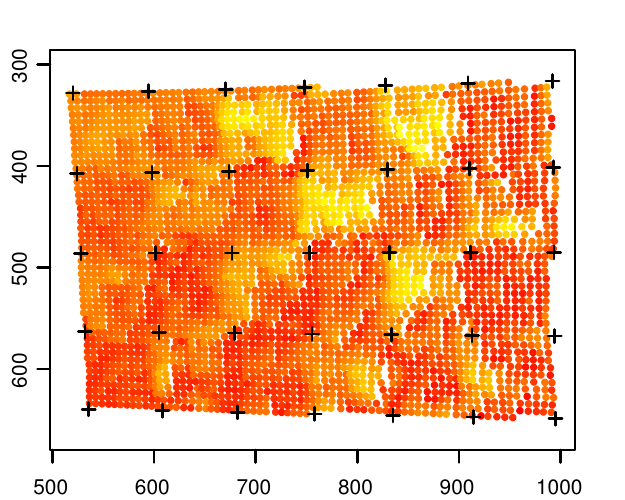}
\caption{\threed\ \tof\ pixels on an example calibration board, projected into an RGB camera in the same (top) and in a different (bottom) system. Each pixel is colour-coded according to the relative total-error (\ref{eqn:tot-err}) in each case (larger errors are shown in yellow). Black crosses mark the detected vertices in the \rgb\ image. The black squares of the board are associated with larger reprojection errors, particularly after reprojecting to a different system (mean error in the bottom example is 10.05px for black squares, and 3.55px for white).}
\label{fig:board-err}
\end{figure}

It is possible to understand these results more fully by examining the distribution
of the total error across individual boards. Figure~\ref{fig:board-err} shows
the distribution for a board reprojected to same/different systems (i.e.~part of the data
from figs.~\ref{fig:pix-intra} \& \ref{fig:pix-inter}). There is a relatively smooth gradient of error
across the board, which is attributable to errors in the fitting of plane $\pi_j$,
and in the estimation of the camera parameters. In addition to these effects, it is clear that the gross errors are correlated with the black squares of the board, which reflect too little of the \tof\ signal. 
For instance, in the bottom example of figure ~\ref{fig:board-err}, the mean reprojection error associated with the black squares is $10.05$ pixels, whereas the white squares are associated with a mean error of $3.55$ pixels. 
The effect is particularly noticeable, as expected from the histograms, when reprojecting to different camera systems.


\subsection{Comparison of homography and similarity transformations}
\label{sec:similarity}

It was argued in section \ref{sec:methods} that the \threed\ homography model is an appropriate way to compensate for miscalibrations of the \tof\ and stereo systems (and indeed it allows the stereo system to be left uncalibrated, if preferred \cite{hansard-2011}). This claim is tested in the following experiments, using data images from two new data sets.

The most obvious alternative to the homography is a rigid motion and scaling; i.e.~a \threed\ \emph{similarity} transformation. Specifically, the $4\times 4$ matrix $\mat{H}$ in (\ref{eqn:homog}) can be replaced by the similarity transformation
\begin{equation}
\mat{S} = 
\begin{pmatrix}
\sigma\,\mat{R} & \mat{t}\\
\mat{0} & 1
\end{pmatrix}
\label{eqn:sim}
\end{equation}
where $\sigma$ is a (positive) scalar, $\mat{R}$ is a $3\times 3$ rotation matrix, and $\mat{t}$ is a $3\times 1$ translation vector. Hence there are only seven degrees of freedom, in contrast to the fifteen of the homography. Geometrically, the similarity model can be interpreted as a rigid transformation between the \tof\ and stereo systems, where the baseline distance of the latter is unknown. 

In principle, a homography should always result in equal or lower error, because it includes the similarity transformation as a special case. 
In particular, a homography can always be written as $\mat{H}=\mat{S}\mat{D}$ where the deformation $\mat{D}$ is composed of an affinity and an elation \cite{hartley-2000}. However, there are two practical issues to consider. Firstly, the quality of the actual estimate in each case; and secondly, the possible danger of over-fitting with the homography.









It was argued in section \ref{sec:methods} that the \threed\ homography model is an appropriate way to compensate for miscalibrations of the \tof\ and stereo systems (and also allows the stereo system to be left uncalibrated, if preferred \cite{hansard-2011}).
In particular the deformation $\mat{D}$ can help to account for locally linear depth bias in the \tof\ camera \cite{foix-2011}. Hence the additional eight degrees of freedom in the homography model allow for an overall approximation to the per-pixel corrections performed by Cui et al. \cite{cui-2013}.

The procedure to estimate the optimal similarity (\ref{eqn:sim}) is analogous to that used to estimate the homography in section \ref{sec:methods}. Instead of DLT, an initial estimate is obtained from the Procrustes algorithm \cite{schoenemann-1970,horn-1988}, using three or more correspondences. The reprojection error is then minimized using the Levenberg-Marquardt procedure, as with the homography. Note that the rotation $\mat{R}$ in (\ref{eqn:sim}) is appropriately parameterized by the Rodrigues formula.

\begin{figure}[!ht]
\mbox{\includegraphics[width=.5\linewidth]{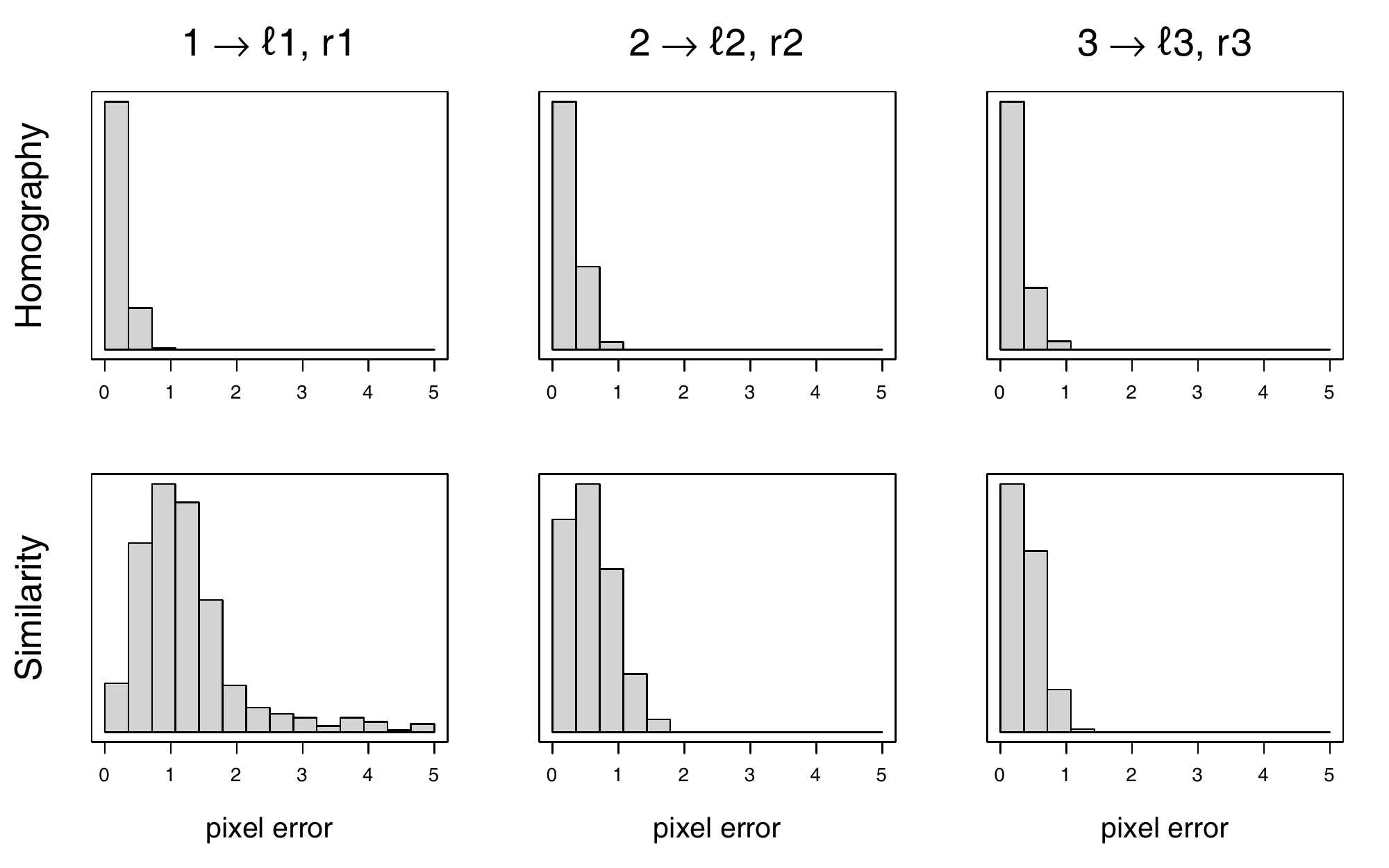} \includegraphics[width=.5\linewidth]{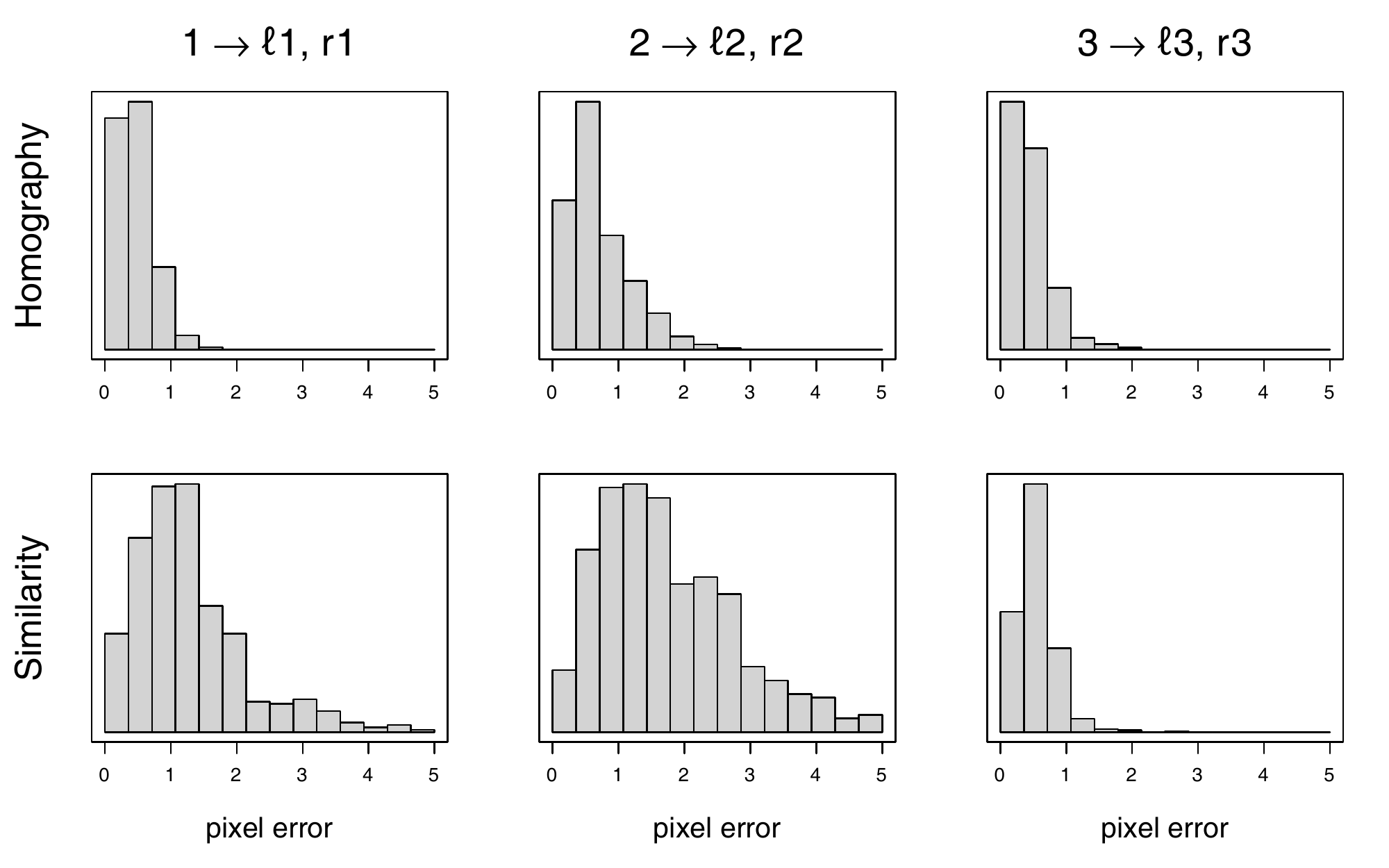}}
\caption{Calibration error (\ref{eqn:cal-err}), measured by projecting the fitted \tof\ points $\mat{Q}^\pi_i$ to the left and right \rgb\ images ($1624\times 1224$) of the \emph{respective} systems $i=1,2,3$. The \tof/stereo\ alignment was computed via a \threed\ homography (top), or a similarity transformation (bottom). It is clear that the former is superior, as described in section \ref{sec:similarity}. The left and right $2\times 3$ blocks represent different capture-sessions, as described in the text.}
\label{fig:hom-sim}
\end{figure}

The results of these experiments are shown in figure \ref{fig:hom-sim}. In every \tofrgb\ system, and in both capture sessions, the homography results in lower error than the similarity. The mean reprojections errors (in pixels) were 0.22 vs.~0.65, and 0.46 vs.~1.06, for the respective capture sessions. The generally higher figures from the second set are due to a broader distribution of board-poses in the evaluation data-set.

These experiments show that the \threed\ homography transformation is a suitable model for \tof/\rgb\ alignment, as argued in section \ref{sec:methods}. It may also be noted that the homography is effectively easier to estimate than the similarity, as there is no need to maintain the orthogonality of $\mat{R}$ during the final minimization.

\subsection{Cross-calibration for \threed\ reconstruction and rendering}
\label{sec:applications}
\begin{figure}[!ht]
\centering
\includegraphics[width=1\linewidth]{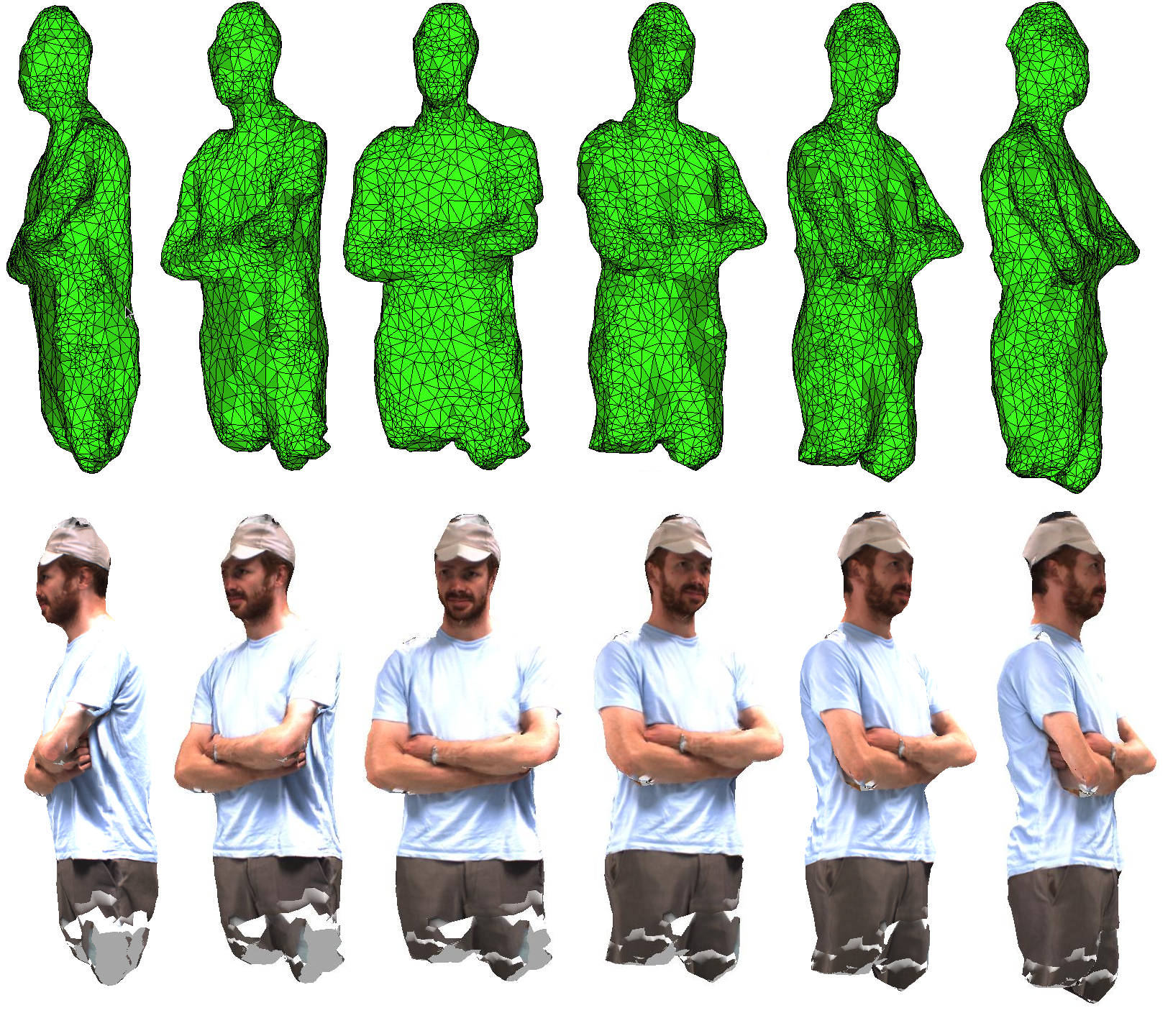}
\caption{
Examples of a segmented $360^\circ$ reconstruction, after Poisson meshing, with and without \rgb\ texture.}
\label{fig:360-reconstruction}
\end{figure}

The proposed cross-calibration methodology also allows a fully-textured \threed\ model to be segmented and rendered. This is because the depth-data are automatically assigned to pixels in the \rgb\ images. Figure~\ref{fig:360-reconstruction} shows reconstruction instances with and without texture from a $360^\circ$ reconstruction, obtained from the cross-calibration of four \tofrgb\ systems. Meshing was performed by the standard Poisson reconstruction method \cite{kazhdan-2006}, followed by simple triangle-based texture mapping.

\begin{figure}[!ht]
\centering
\includegraphics[width=5.5cm,height = 9cm]{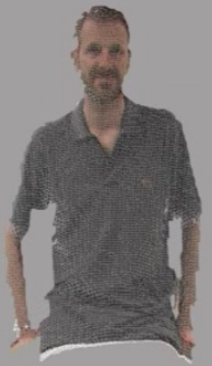}
\includegraphics[width=5.5cm,height = 9cm]{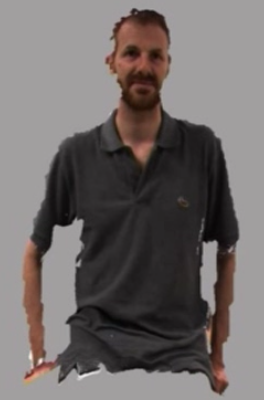}\\
\includegraphics[width=5.5cm,height = 9cm]{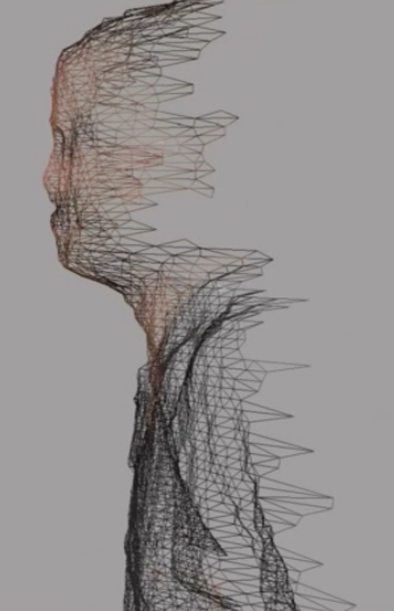}
\includegraphics[width=5.5cm,height = 9cm]{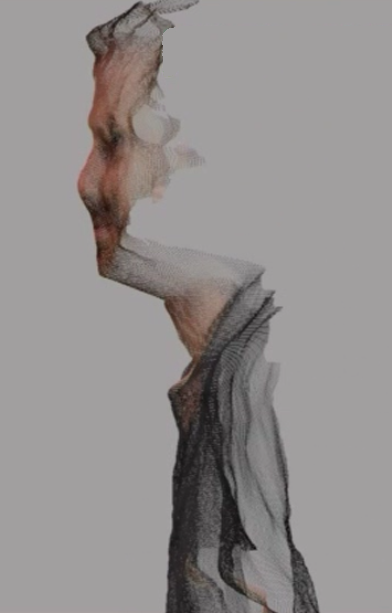}
\caption{
Cross-calibration for \threed\ reconstruction. \emph{Left:} The raw mesh, with textured edges, obtained directly from our \tofrgb\ cross-calibration method. \emph{Right:} The textured high-density point cloud, obtained after applying the \tof{}-stereo fusion method of~\cite{gandhi-2012} to the cross-calibration results.}
\label{fig:sparse_vs_dense_recon}
\end{figure}

More dense reconstruction can be achieved by exploiting the full resolution of stereo cameras, as described elsewhere \cite{gandhi-2012}. Figure~\ref{fig:sparse_vs_dense_recon} shows the difference between the raw reconstruction obtained from the cross-calibration of a \tofrgb\ system, and that obtained after using the \tof\ data in a subsequent stereo algorithm~\cite{gandhi-2012}. A high-resolution depth map is produced, making full use of the high-resolution \rgb\ cameras.

\color{black}

\section{Conclusion}
\label{sec:conclusion}

It has been shown that \threed\ projective transformations can be used to cross-calibrate a system of \tof\ and stereo reconstructions.
A practical method for computing these transformations, based on geometric principles, has been introduced. This calibration procedure has been extended to a nine-camera network of  six \rgb\ and  three \tof\ cameras, and evaluated in detail. 

A clear distinction has been made between \emph{calibration error}, which is due to imperfect camera and image models, versus \emph{total error}, which incorporates \tof-specific noise and biases. It has been shown that these errors can be separated geometrically, and used to characterize the performance of a \tof/\rgb\ camera network.
The overall performance of the system has been visualized in segmented 360$^\circ$ \threed\ reconstructions. This type of reconstruction is challenging to compute, and shows that the system is more than adequate as a basis for scene-segmentation tasks.

The accuracy of the method presented here is somewhat limited by the relatively poor localization of points, even when spatially interpolated, in the \tof\ images. The detection of standard chequerboard patterns, in \tof\ images, has been discussed elsewhere \cite{hansard-2014}. The design of more convenient calibration patterns, for \tof\ cameras, would be a worthwhile direction for future research.
Meanwhile, in mitigation, the spatial resolution of \tof\ cameras will continue to increase. 

Future work should also consider the distribution of range-errors in \threed, and how this can be used to design custom meshing and surface reconstruction algorithms for time-of-flight data. This, in conjunction with the \rgb\ textures provided by the present method, would lead to a comprehensive approach to \threed\ reconstruction, rendering and scene-understanding.

\bibliographystyle{elsarticle-num}

\end{document}